\documentclass[preprint,12pt]{elsarticle}%
\usepackage{graphicx}
\usepackage{float}
\usepackage{url}
\usepackage{amsmath}
\usepackage{amsfonts}
\usepackage{amssymb}
\usepackage[letterpaper]{geometry}%
\usepackage[ruled,vlined]{algorithm2e}
\usepackage{comment}
\usepackage{subfigure}
\providecommand{\U}[1]{\protect\rule{.1in}{.1in}}
\begin{document}

%

\begin{frontmatter}

\title{Automatic Model Selection for Neural Networks}

\author{David Laredo$^{1}$, Yulin Qin$^{1}$, Oliver Sch\"utze$^{2}$ and Jian-Qiao Sun$^{1}$}
\address{
$^{1}$Department of Mechanical Engineering, School of Engineering\\
University of California, Merced, CA 95343, USA\\
$^{2}$Department of Computer Science, CINVESTAV, Mexico City, Mexico\\
Corresponding author: Jian-Qiao Sun. Email: jqsun@ucmerced.edu}

\begin{abstract}
Neural networks and deep learning are changing the way that artificial intelligence is being done. Efficiently choosing a suitable network architecture and fine tune its hyper-parameters for a specific dataset is a time-consuming task given the staggering number of possible alternatives. In this paper, we address the problem of model selection by means of a fully automated framework for efficiently selecting a neural network model for a given task: classification or regression. The algorithm, named Automatic Model Selection, is a modified micro-genetic algorithm that automatically and efficiently finds the most suitable neural network model for a given dataset. The main contributions of this method are: a simple list based encoding for neural networks as genotypes in an evolutionary algorithm, new crossover and mutation operators, the introduction of a fitness function that considers both, the accuracy of the model and its complexity and a method to measure the similarity between two neural networks. AMS is evaluated on two different datasets. By comparing some models obtained with AMS to state-of-the-art models for each dataset we show that AMS can automatically find efficient neural network models. Furthermore, AMS is computationally efficient and can make use of distributed computing paradigms to further boost its performance. 
\end{abstract}


\end{frontmatter}


\section{Introduction}

Machine learning studies algorithms that improve themselves through experience. Given the large amount of data currently available in many fields such as engineering, bio-medical, finance, \textit{etc.} and the increasingly more computing power available, machine learning is now practiced by people with very diverse backgrounds. More users of machine learning tools are non-experts who require off-the-shelf solutions. Automated Machine Learning (AutoML) is the field of machine learning devoted to developing algorithms and solutions to enable people with limited machine learning background knowledge to use machine learning models easily. Tools like WEKA \cite{Hall2009}, PyBrain \cite{Schaul2010} and MLLib \cite{mlib2017} follow this paradigm. Nevertheless, the user still needs to make some choices, which may not be obvious or intuitive in selecting a learning algorithm, hyper-parameters, features, \textit{etc.} and thus leads to the selection of non-optimal models.

Recently, deep learning models such as CNN, RNN and Deep NN have gained a lot of attention due to their improved efficiency on complex learning problems and their flexibility and generality for solving a large number of problems including regression, classification, natural language processing, recommendation systems, \textit{etc}. Furthermore, there are many software libraries which make their implementation easier. TensorFlow \cite{TensorFlow2015}, Keras \cite{keras2015}, Caffe \cite{caffe2014} and CNTK \cite{cntk2016} are some examples of such libraries. Despite the availability of such libraries and tools, the tasks of picking the right neural network model and its hyper-parameters are usually complex and iterative in nature, specially among non-computer scientists.

Usually, the process of selecting a suitable machine learning model for a particular problem is done in an iterative manner. First, an input dataset must be transformed from a domain specific format to features which are predictive of the field of interest. Once the features have been engineered, users must pick a learning setting which is appropriate to their problem, e.g. regression, classification or recommendation. Next, users must pick an appropriate model, such as support vector machines (SVM), logistic regression or any flavor of neural networks (NNs). Each model family has a number of hyper-parameters, such as regularization degree, learning rate, number of neurons, \textit{etc.} and each of these must be tuned to achieve optimal results. Finally, users must pick a software package that can train their model, configure one or more machines to execute the training and evaluate the model's quality. It can be challenging to make the right choice when facing so many degrees of freedom, leading many users to select a model based on intuition or randomness and/or leave hyper-parameters set to default. This approach will usually yield sub-optimal results.

This suggests a natural challenge for machine learning: given a dataset, to automatically and simultaneously choose a learning algorithm and set its hyper-parameters to optimize performance. As mentioned in \cite{Hall2009}, the combined space of learning algorithm and hyper-paremeters is very challenging to search: the response function is noisy and the space is high dimensional involving both, categorical and continuous choices and containing hierarchical dependencies, e.g. hyper-parameters of the algorithm are only meaningful if that algorithm is chosen. Thus, identifying a high quality model is typically costly in the sense that it entails a lot of computational effort and time-consuming.

To address this challenge, we propose Automatic Model Selection (AMS), a flexible and scalable method to automate the process of selecting artificial neural network models. The key contributions of the method are: 1) a simple, list based encoding of neural networks as genotypes for evolutionary computation algorithms, 2) new crossover and mutation operators to generate valid neural networks models from an evolutionary algorithm, 3) the introduction of a fitness function that considers both, the accuracy of the model and its complexity and 4) a method for measuring the similarity between two neural networks. All these components together form a new method based on an evolutionary algorithm, which we call AMS, and that can be used to find an optimal neural network architecture for a given dataset.

The remainder of this paper is organized as follows: Section \ref{sec:model_selection} formally introduces the model selection problem. The related work is briefly reviewed in Section \ref{sec:literature_review}. The AMS algorithm and all of its components are described in detail in Section \ref{sec:auto_nn}, experiments to test the algorithm and comparison against other state of the art methods are presented in Section \ref{sec:evaluation}. Conclusions and future work are discussed in Section \ref{sec:conclusions}.


\section{Problem Statement}
\label{sec:model_selection}

In this section we mathematically define the general neural network architecture search problem. Consider a dataset $\mathcal{D}$ made up of training points $d_i = (\mathbf{x}_i, \mathbf{y}_i) \in \mathbf{X} \times \mathbf{Y}$ where $\mathbf{X}$ is the set of data points and $\mathbf{Y}$ is the set of labels. Furthermore, we split the dataset into training set $D_t$, cross validation set $\mathcal{D}_v$, and test set $\mathcal{D}_p$. Given a neural network architecture search space $\mathcal{H}$, the performance of a neural network architecture $\phi \in \mathcal{H}$ trained on $\mathcal{D}_t$ and validated with  $\mathcal{D}_v$ is defined as:
\begin{equation}
p = {Perf}(\phi(\mathcal{D}_t), \mathcal{D}_v),
\label{eq:nn_cost}
\end{equation}
where ${Perf}(.)$ is a measurement of the generalization error attained by the learning algorithm $\phi(.)$ on the validation set $\mathcal{D}_v$. Common error indicators are accuracy error, precision error and mean squared error. Their definitions along with some other common error indicators are presented in Table \ref{table:performance_metrics}. 

\begin{table}[H]
\begin{center}
\begin{tabular}{| c | c | c |}
\hline
Indicator name & Application & Definition\\[2ex]
\hline
Mean Squared Error & Regression & $E_{MSE} = \frac{1}{2}\sum\limits_{i=1}^{n} \left( \mathbf{\hat{y}} - \mathbf{y} \right)^2$\\[2ex]
Accuracy & Classification & $E_{A} = \frac{tp+tn}{tp+tn + fp + fn}$ \\[2ex]
Precision & Classification & $E_{P} = \frac{tp}{tp+fp}$ \\[2ex]
Recall & Classification & $E_{R} = \frac{tp}{tp+fn}$\\[2ex]
F1 & Classification & $F_{1} = 2 \frac{E_{P} E_{R}}{E_{P} + E_{R}}$\\[2ex]
\hline
\end{tabular}
\end{center}
\caption{Common performance metrics for neural networks. $\mathbf{\hat{y}}$ represents the predicted value of the model for a sample $\mathbf{x}$. $tp$ stands for true positives count. $tn$ stands for true negatives count. $fp$ stands for false positives count. $fn$ stands for false negatives count.}
\label{table:performance_metrics}
\end{table}

Finding a neural network $\phi^* \in \mathcal{H}$ that achieves a good performance has been explored in \cite{AutoKeras2018,Real2018} among others. While this task alone is challenging, usually the efficiency of $\phi^*$ is not measured. Indeed, it turns out that there can be several candidate models that can attain similar performances with improved efficiency. By efficiency we mean, in practical terms, how fast is to train $\phi^*$ as compared to other possible solutions. 

We aim at achieving neural network models $\phi^*$ that not only exhibit good performance on $\mathcal{D}$ as measured by $p$, but also achieve such performance by using a simple structure, which directly translates to improved efficiency of the model. To measure the complexity of the architecture, we make use of the number of trainable parameters $w(\phi)$ of the neural network. We will also refer to $w(\phi)$ as the ``size'' of the neural network.

The problem of finding a neural network $\phi$ that achieves a good performance on the dataset $\mathcal{D}$ while using a simple model can be mathematically stated as the following multi-objective optimization problem:

\begin{equation}
	\begin{aligned}
	\underset{\phi \in \mathcal{H}}{\text{min}}
	& \quad (p(\phi), w(\phi))\\
	\end{aligned}
	\label{eq:model_selection_eq}
\end{equation}

In this paper we develop an algorithm to efficiently solve Problem (\ref{eq:model_selection_eq}).


\section{Related Work}
\label{sec:literature_review}

Automated machine learning has been of research interest since the uprising of deep learning. This is no surprise since selecting an effective combination of algorithm and hyper-parameter values is currently a challenging task requiring both deep machine learning knowledge and repeated trials. This is not only beyond the capability of layman users with limited computing expertise, but also often a non-trivial task even for machine learning experts \cite{sparks2015}. 

Until recently most state-of-the-art neural network architectures have been manually designed by human experts. To make the process easier and faster, researchers have looked into automated methods. These methods intend to find, within a pre-specified resource limit in terms of time, number of algorithms and/or combinations of hyper-parameter values, an effective algorithm and/or combination of hyper-parameter values that maximize the accuracy measure on the given machine learning problem and data set. Using an automated machine learning, the machine learning practitioner can skip the manual and iterative process of selecting an efficient combination of hyper-parameter values and neural network model, which is labor intensive and requires a high skill set in machine learning.

In the context of deep learning, neural architecture search (NAS), which aims at searching for the best neural network architecture for the given learning task and dataset, has become an effective computational tool in AutoML. Unfortunately, the existing NAS algorithms are usually computationally expensive where its time complexity can easily scale to $O(nt)$ where $n$ is the number of neural architectures evaluated, and $t$ is the average time consumption for each of the $n$ neural networks. Many NAS approaches such as deep reinforcement learning  \cite{Zoph2016,Baker2016,Zhong2017} and evolutionary algorithms \cite{Liu2018,Liang2017,Angeline1994,Suganuma2017} require a large $n$ to reach good performance. Other approaches, including Bayesian Optimization \cite{Thornton2016,Brochu2010} and Sequential Model Based Optimization (SMBO) \cite{Hutter2011,Feurer2015}, are often as expensive as NAS while being more limited to the kind of models they can explore.

In the recent years, a number of tools have been made available for users to automate the model selection and/or hyper-parameter tuning. In the following, we present a brief description of some of them.

\subsection{Auto-Keras}

Auto-Keras \cite{AutoKeras2018} is a method to automatically generate model architectures. It defines an edit-distance kernel to measure the difficulty of transferring the current model to a  new one. This kernel makes it possible to search the model structures in a tree-structured space constructed from the network morphism. Bayesian optimization is used along with the kernel to optimize the tree-structure of the model. The consistency of the input and output shapes is guaranteed throughout the process. The use of Auto-Keras does not require an extensive knowledge of machine learning, making it very accessible for starters. However, training the model is very time-consuming since a new model must be trained from scratch every time. Furthermore,  acquiring the parameters in the middle steps is also computational expensive. An example on MNIST dataset may take several hours to converge (depending on the used equipment).

\subsection{AutoML Vision}

AutoML \cite{AutoML2017} uses evolutionary algorithms to perform image classification. Without any restriction on the search space such as network depth, skip connections, \textit{etc}., the algorithm starts from a simple model without convolutions and iteratively evolves the model into more a complex one. A massively-parallel and lock-free infrastructure is designed, and many computers may be used to search the large solution space. Communication between different nodes in the network is handled by using a shared file system that keeps track of the population. Among the main disadvantages of AutoML are that it requires extremely high computational power. Furthermore, since the candidate models start from a very simple structure poor performing models are likely to be obtained as a solution.

\subsection{Auto-sklearn}

Auto-sklearn \cite{Feurer2015} is a system designed to help machine learning users by automatically searching through the joint space of sklearn's learning algorithms and their respective hyper-parameter settings to maximize performance using a state-of-the-art Bayesian optimization method. Auto-sklearn addresses the model selection problem by treating all of sklearn algorithms as a single, highly parametric machine learning framework, and using Bayesian optimization to find an optimal instance for a given dataset. Auto-sklearn also natively supports parallel runs on a single machine to find good configurations faster and save the $n$ best configurations of each run instead of just the single best. Nevertheless, and to the best of our knowledge, Auto-sklearn does not provide support for neural networks and it does not take into consideration the complexity of the proposed models for assessing their optimality.


\section{Proposed Evolutionary Algorithm for Model Selection}
\label{sec:auto_nn}

While there are a number of methods for automatic model selection and hyper-parameter tuning, many of them do not provide support for some of the most sophisticated deep learning architectures. For instance, Auto-sklearn does not provide good support for deep learning methods, support for distributed computing is also very limited. On the other hand Auto-Keras and AutoML Vision do provide support for deep learning methods, but they do not consider the model's complexity when assessing its overall performance. Indeed, Auto-Keras and AutoML require clusters of or hours of computing time to yield models with good accuracy. Furthermore, Auto-Keras and AutoML do not provide good support for regression problems. 

We propose an efficient method that, for a given dataset $\mathcal{D}$, will automatically find a neural network model that attains high performance while being computationally efficient. The proposed model is capable of performing inference tasks for both classification and regression problems. Furthermore, the proposed system is scalable and easy to use in distributed computing environments, allowing it to be usable for large datasets and complex models. For this task, we make use of Ray \cite{Moritz2017}, a distributed framework designed with large scale distributed machine learning in mind.

Our method provides support for three of the major neural networks architectures, namely multi-layer percepetrons (MLPs) \cite{Engelbrecht2007}, convolutional neural networks (CNNs) \cite{imagenet_cvpr09} and recurrent neural networks (RNNs) \cite{Lipton15}. Our method can construct models of any of these architectures by stacking together a \textit{valid} combination of any of the four following layers: fully connected layers, recurrent layers, convolutional layers and pooling layers. Our method does not only build neural networks for the aforementioned architectures, but also tunes some of the hyper-parameters such as the number of neurons at each layer, the activation function to use or the dropout rates for each layer. Support for skip connections is left for future work.

We say that a neural network architecture is \textit{valid} if it complies with the following set of rules, which we derived empirically from our practice in the field:

\begin{itemize}
\item A fully connected layer can only be followed by another fully connected layer.
\item A convolutional layer can be followed by a pooling layer, a recurrent layer, a fully connected layer or another convolutional layer.
\item A recurrent layer can be followed by another recurrent layer or a fully connected layer.
\item The first layer is user defined according to the type of architecture chosen (MLP, CNN or RNN).
\item The last layer is always a fully connected layer with either a softmax activation function for classification or a linear activation function for regression problems.
\end{itemize}

\subsection{Automatic Model Selection (AMS)}

The key idea of our method is to develop an evolutionary algorithm (EA) which is capable of evolving different neural network architectures to find a suitable model for a given dataset $\mathcal{D}$, while being computationally efficient. EAs are chosen for this work since, contrary to the more classical optimization techniques, they do not make any assumptions about the problem, treating it as a black box that merely provides a measure of quality for given a candidate solution. Furthermore, they do not require the gradient, which is impossible to obtain for a neural network $\phi$ when searching for optimal solutions.

In the following, we describe the very basics of evolutionary algorithms as an introduction for the reader. Further reading can be found in \cite{Engelbrecht2007,Ebehart2007,Sumathi2010}. 

Every evolutionary algorithm consists of a population of individuals which are potential solutions to the optimization problem such as the one described by the fitness function in Equation (\ref{eq:fitness_function_scaled}). Each individual in the population is a neural network model. Every individual has a specific genotype or encoding, in the evolutionary algorithm domain, that represents a solution to the given problem while the actual representation of the individual, in the specific application domain, is often referred as the phenotype. For the current application, the phenotype represents the neural network architecture while the genotype is represented by a list of lists. When assessing the quality of an individual, EA makes use of a so-called fitness function, which indicates how every individual in the population performs with respect to a certain performance indicator, establishing thus an absolute order among the various solutions and a way of fairly comparing them against each other. 

New generations of solutions are created iteratively by using crossover and mutation operators. The crossover operator is an evolutionary operator used to combine the information of two parents to generate a new offspring while the mutation operator is used to maintain genetic diversity from one generation of the population to the next.

The basic template for an evolutionary algorithm is described in Algorithm \ref{algorithm:generic_ea}.

\begin{algorithm}[H]
\SetAlgoLined
\KwData{None}
\SetKwInOut{Input}{Input}\SetKwInOut{Output}{Output}

\Input{An objective function $f(\mathbf{x})$}
\Output{A vector $\mathbf{x^*}$ such that $f(\mathbf{x}^*)$ is a local minimum.}

\BlankLine

Let $t = 0$ be the generation counter.
Create and initialize an $n_x$-dimensional population, $\mathcal{C}(0)$, consisting of $n$ individuals.

\While{Stopping condition not true.}{
	Evaluate the fitness, $f(\mathbf{x}_i(t))$, of each individual, $\mathbf{x}_i(t)$ in the population.\\
	Perform reproduction to create offspring.\\
	Select the new population, $\mathcal{C}(t+1)$.\\
	Advance to the new generation, i.e. $t = t +1$.
}

\caption{Basic evolutionary algorithm.}
\label{algorithm:generic_ea}
\end{algorithm}

One of the major drawbacks of EAs is the time penalty involved in evaluating the fitness function. If the computation of the fitness function is computationally expensive, as in this case, then using any variant of EA may be very computationally expensive and in some instances unfeasible. Micro-genetic algorithms \cite{Krishnakumar1989} are one variant of GAs whose main advantage is the use of small populations, for example, less than 10 individuals per population, in contrast to some other EAs like the genetic algorithms (GAs), evolutionary strategies (ES) and genetic programming (GP) \cite{Engelbrecht2007}. Since computational efficiency is one of our main concerns for this work, we will follow the general principles of micro-GA in order to reduce the computational burden of the proposed method. 

The pseudocode for our proposed method is described in Algorithm \ref{algorithm:nn_ea}. Let $\rho_c$ and $\rho_m$ be the crossover and mutation probabilities, respectively. Let $\gamma_g$ be the maximum number of allowed generations and $\gamma_r$ the maximum number of repetitions for the micro-GA. Finally, let $\mathcal{B}$ be an archive for storing the best architectures found at every run of the micro-GA. Our algorithm Automatic Model Selection (AMS) is stated in Algorithm \ref{algorithm:nn_ea}. In the following sections, we describe in detail each one of the major components of the AMS algorithm.

\begin{algorithm}[H]
\SetAlgoLined
\KwData{Training/cross-validation dataset $\mathcal{D}_t$, $\mathcal{D}_v$.}
\SetKwInOut{Input}{Input}\SetKwInOut{Output}{Output}

\Input{Algorithm's hyper-parameters. See Table \ref{table:MNIST_params} for an example. }
\Output{The most suitable neural network $\phi^*$ according to the user preferences.}

\BlankLine

Let $t_e = 0$ be the experiments counter
Create and initialize an $n_x$-dimensional population, $\mathcal{C}(0)$, consisting of $n$ individuals.

\While{$t_e < \gamma_r$}{
	Let $t_g = 0$ be the generation counter.\\
	Create and initialize an initial population $\mathcal{C}(0)$, consisting of $n$ individuals, where $n <= 10$. See section \ref{sec:valid_models}.\\
	\While{$t_g < \gamma_g$ or nominal convergence not reached}{
		Check for nominal convergence in $\mathcal{C}(t)$. See section \ref{sec:convergence}.\\
		Evaluate the cost, $c(\phi)$ of each candidate model $f$. See section \ref{sec:fitness_function}.\\
		Identify best and worst models in $\mathcal{C}(t)$.\\
		Replace worst model in $\mathcal{C}(t)$ with best from $\mathcal{C}(t-1)$.\\
		Perform selection. See section \ref{sec:selection}.\\
		Perform crossover of models in $\mathcal{C}(t)$ with $\rho_c=1$. Let $\mathcal{O}(t)$ be the offspring population. See section \ref{sec:crossover}.\\
		For each model in $\mathcal{O}(t)$ perform mutation with $\rho_m$ probability. See section \ref{sec:mutation}.\\
		Make $\mathcal{C}(t+1) = \mathcal{O}(t)$.\\
		$t_g = t_g + 1$.\\
	}
	Append best solution from previous run to $\mathcal{B}$.\\
	$t_e = t_e + 1$.\\
}
Normalize the cost for each model in the archive $\mathcal{B}$. See section \ref{sec:fitness_function}.\\
Final Solution is best existing solution in $\mathcal{B}$.\\

\caption{Automatic model selection algorithm.}
\label{algorithm:nn_ea}
\end{algorithm}

\subsection{The Fitness Function}
\label{sec:fitness_function}

To establish a ranking among the different tested architectures, a suitable cost or fitness function is required. While Equation (\ref{eq:model_selection_eq}) can be used as the cost function, this would give rise to a multi-objective optimization problem (MOP). We leave this approach for a future revision of this work and instead make use of scalarization to transform the MOP into a single-objective optimization problem (SOP). The scalarization approach taken here is the well known weighted sum method \cite{Hillermeier2001}. Equation (\ref{eq:model_selection_eq}) is restated as
\begin{equation}
\begin{aligned}
	\underset{\phi \in \mathcal{H}}{\text{min}}
	& \quad (1-\alpha)p(\phi) + \alpha w(\phi).\\
\end{aligned}
\label{eq:model_selection_scalar}
\end{equation}

The cost function associated with Equation (\ref{eq:model_selection_scalar}) is
\begin{equation}
c(\phi) = (1-\alpha)p(\phi) + \alpha w(\phi), 
\label{eq:fitness_function_scalar}
\end{equation}
where $\alpha \in \left[0,1\right]$ is a scaling factor biasing the total cost towards the size of the network or its performance. Equation (\ref{eq:model_selection_scalar}) measures the cost in terms of performance and size of a given neural network $\phi$.

Note that making an accurate assessment of the inference performance, $p(\phi)$, of a neural network involves training $\phi$ for a large number of epochs. Since the training process usually involves thousands of computations, training every candidate solution and then assessing its performance becomes unfeasible. Instead, we relax the training process for each of the candidate models by using a \textit{partial training strategy} which, in short, is training the model for a very small number of epochs, for example, only a few tens of them. This approach was been successfully tested in \cite{Laredo2019a}. Even though the models are just partially trained, a clear trend in terms of whether a model is promising or not can be clearly observed.

Computing the fitness of individuals using the current definitions of $p(\phi)$ and $w(\phi)$ present a challenge because the performance indicator $p(\phi)$ and the number of trainable weights $w(\phi)$ are on entirely different scales. While $w(\phi)$ can range from a few hundreds up to several millions, the range of $p(\phi)$ depends on the type of scoring function used. Table \ref{table:neural_network_common_ranges} presents some common ranges for $p(\phi)$. 

\begin{table}[H]
\begin{center}
\scalebox{1}{
\begin{tabular}{| c | c | c |}
\hline
${Perf}(.)$ & Range & Common range \\
\hline
Accuracy & $\left[ 0, 1 \right]$ & $\left[ 0, 1 \right]$\\
Precision & $\left[ 0, 1 \right]$ & $\left[ 0, 1 \right]$\\
Recall & $\left[ 0, 1 \right]$ & $\left[ 0, 1 \right]$\\
MSE & $\left[ 0, +\infty \right)$ & $\left[ 0, 10^4 \right]$\\
RMSE & $\left[ 0, +\infty \right)$ & $\left[ 0, 10^2 \right]$\\
\hline
\end{tabular}
}
\end{center}
\caption{Common ranges for some neural network performance indicators.}
\label{table:neural_network_common_ranges}
\end{table}

For the present application, it is necessary that the range of $p(\phi)$ is consistent independent of the type of score ${Perf}(.)$. Thus, we normalize the values of $p(\phi)$ to be within the range $[ 0,1]$. The normalization is done as follows. Assume that a population $\mathcal{C}$ has $n$ individuals. Let $\mathbf{p} = \left[ p_0(f), p_1(f), \ldots, p_n(f) \right]$ be the vector whose components are the scores $p_i$ for the $i^{th}$ element in the population. Then $\mathbf{p}^* = \mathbf{p}/norm_2(\mathbf{p})$ is the vector with the normalized values of $\mathbf{p}$, hence $p^*_i \in \left[ 0, 1 \right]$ for any score ${Perf}(.)$.

Now we focus on $w(\phi)$. For the sake of simplicity, let us just consider the case of the MLP class of neural networks since this is usually the model where $w(\phi)$ is larger. Let $\mathcal{A}$ be the maximum number of possible layers for any model. For this work, we limit $\mathcal{A} = 64$ since this is a reasonable number for most mainstream deep learning models. Table \ref{table:neural_network_array_details} shows that the maximum number of neurons at any layer is set to be 1024. Thus, the maximum number of $w(\phi)$ for any given model is $\mathcal{W} = 2^{26}$. Furthermore, we want neural networks that are similar to have the same score $w(\phi)$. Therefore, we replace the last 3 digits of the $w(\phi)$ with $0's$, and let $w^+(\phi)$ be this new size score. Finally, considering that $p^*(\phi) \in \left[ 0, 1 \right]$, we make $w^*(\phi) = log(w^+(\phi))$, therefore $\mathcal{W}^* \approx  log(2)*26$. Hence, $w^*(\phi) \in \left[0, 7.8\right]$ which is in the same order of magnitude as $p^*(\phi)$.

Thus, we rewrite Equation (\ref{eq:fitness_function_scalar}) as
\begin{equation}
c(\phi) = 10(1-\alpha)p^*(\phi) + \alpha w^*(\phi), 
\label{eq:fitness_function_scaled}
\end{equation}
where we multiply $p^*(\phi)$ by a factor of $10$ to make the scaling similar to that of $w^*(\phi)$. As can be observed, Equation (\ref{eq:fitness_function_scaled}) is now properly scaled. Therefore, it is a suitable choice as the fitness function for assessing the performance of a neural network model while also considering its size.

\subsection{Neural Networks as Lists}
\label{sec:encoding_nn}

In order to perform the optimization of neural network architectures, a suitable encoding for the neural networks is needed. A good encoding has to be flexible enough to represent neural network architectures of variable length while also making it easy to verify the \textit{validity} of the candidate neural network architecture. 

Array based encodings are quite popular for numerical problems. They often use a fixed-length genotype which is not suitable for representing neural network architectures. While it is possible to use an array based representation for encoding a neural network, this would require the use of very large arrays. Furthermore, verifying the validity of the encoded neural network is hard to accomplish. Tree-based representation as those used in genetic programming \cite{Engelbrecht2007} enables more flexibility when it comes to the length of the genotype. Imposing constraints for building a valid neural network requires traversing the entire tree or making use of complex data structures every time a new layer is to be stacked in the model. 

\begin{figure}[H]
\centering
\includegraphics[scale=0.4]{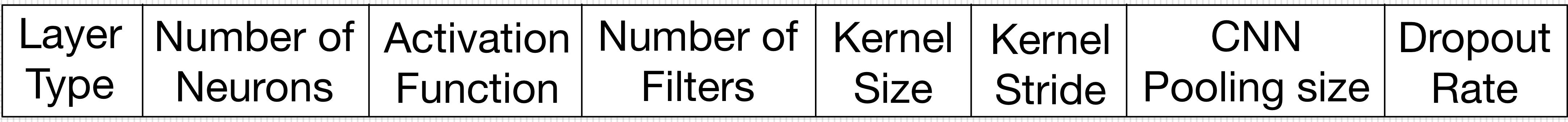}
\caption{Visual representation of a neural network layer as an array.}
\label{fig:neural_network_array}
\end{figure}

In this work, we introduce a list-based encoding. In this new list-based encoding, neural network models are represented as a list of arrays, where the length of the list can be arbitrary. Each array within the list represents the details of a neural network layer as described in Table \ref{table:neural_network_array_details}. A visual depiction of the array is presented in Figure \ref{fig:neural_network_array}.

\begin{table}[H]
\begin{center}
\resizebox{\columnwidth}{!}{%
\begin{tabular}{| c | c | c | c | c |}
\hline
Cell name & Data Type & Represents & Applicable to & Values\\
\hline
Layer type & Integer & The type of layer. See table \ref{table:neural_network_building_rules} & MLP/RNN/CNN & $x \in \left\lbrace 1, \ldots, 5 \right\rbrace$\\
Neuron number & Integer & Number of neurons/units in the layer & MLP/RNN & $8*x$ where $x \in \left\lbrace 1, \ldots, 128 \right\rbrace$ \\
Activation function & Integer &Type of activation function. See table \ref{table:index_to_activation_functions} & MLP/RNN/CNN & $x \in \left\lbrace 1, \ldots, 4 \right\rbrace$\\
Number of filters & Integer & Number of convolution filters & CNN & $8*x$ where $x \in \left\lbrace 1, \ldots 64 \right\rbrace$\\
Kernel size & Integer & Size of the convolution kernel & CNN & $3^x$ where $x \in \left\lbrace 1, \ldots, 6 \right\rbrace$\\
Kernel stride & Integer & Stride used for convolutions & CNN & $x \in \left\lbrace 1, \ldots, 6 \right\rbrace$\\
Pooling size & Integer & Size for the pooling operator & CNN & $2^x$ where $x \in \left\lbrace 1, \ldots 6 \right\rbrace$\\
Dropout rate & Float & Dropout rate applied to the layer & MLP/RNN/CNN & $x \in \left[0,1\right]$\\
\hline
\end{tabular}}
\end{center}
\caption{Details of the representation of a neural network layer as an array.}
\label{table:neural_network_array_details}
\end{table}

\begin{table}[H]
\begin{center}
\begin{tabular}{| c | c | c |}
\hline
Layer type & Layer name & Can be followed by \\
\hline
1 & Fully connected & $\left[ 1, 5 \right]$ \\
2 & Convolutional & $\left[ 1, 2, 3, 5 \right]$ \\
3 & Pooling & $\left[ 1, 2 \right]$\\
4 & Recurrent & $\left[ 1, 4 \right]$\\
5 & Dropout & $\left[ 1, 2, 4 \right]$\\
\hline
\end{tabular}
\end{center}
\caption{Neural network stacking/building rules.}
\label{table:neural_network_building_rules}
\end{table}

\begin{table}[H]
\begin{center}
\begin{tabular}{| c | c |}
\hline
Index & Activation function \\
\hline
0 & Sigmoid \\
1 & Hyperbolic tangent \\
2 & ReLU \\
3 & Softmax \\
4 & Linear \\
\hline
\end{tabular}
\end{center}
\caption{Available activation functions.}
\label{table:index_to_activation_functions}
\end{table}

The proposed representation is capable of handling different types of neural network architectures. In principle, the representation can handle multi-layer perceptrons (MLPs), convolutional neural networks (CNNs) and recurrent neural networks (RNNs). For any given neural network layer, the array will only contain values for the entries that are applicable to the layer, values for other types of layers are set to 0.

Let us illustrate the proposed encoding with an example. The following example considers an MLP. Layer type, Number of neurons, Activation function and Dropout rate entries are applicable values. Consider $S_e$ as a model made up of several stacked layers as shown in Figure \ref{fig:neural_network_array}.
The neural network representation of the presented model is shown in Table \ref{table:neural_network_model}.

\begin{align*}
S_e = \left[ \left[1, 264, 2, 0, 0, 0, 0, 0 \right], \left[5, 0, 0, 0, 0, 0, 0, 0.65 \right], \right. \\
\left. \left[1, 464, 2, 0, 0, 0, 0, 0 \right], \left[5, 0, 0, 0, 0, 0, 0, 0.35 \right], \right. \\
\left. \left[1, 872, 2, 0, 0, 0, 0, 0 \right], \left[1, 10, 3, 0, 0, 0, 0, 0 \right] \right]
\end{align*}

\begin{table}[H]
\begin{center}
\begin{tabular}{| c | c | c | c |}
\hline
Layer Type & Neurons & Activation Function & Dropout Ratio \\
\hline
Fully Connected & 264 & ReLU & n/a \\
Dropout & n/a & n/a & 0.65 \\
Fully Connected & 464 & ReLU & n/a\\
Dropout & n/a & n/a & 0.35\\
Fully Connected & 872 & ReLU & n/a\\
Fully Connected & 10 & Softmax & n/a\\
\hline
\end{tabular}
\end{center}
\caption{Neural network model.}
\label{table:neural_network_model}
\end{table}

Encoding the neural network as a list of arrays has two advantages. First, the number of layers that can be stacked is, in principle, arbitrary. Second, the validity of an architecture can be verified every time a new layer is to be stacked to the model. This is due to the fact that in order to stack a layer to the model, one only needs to check the compatibility between the previous and next layers. The ability of stacking layers dynamically and verifying its correctness when a new layer is stacked allows for a powerful representation. We can build several kinds of neural networks such as fully connected, convolutional and recursive. The rules for stacking layers together are described in Table \ref{table:neural_network_building_rules}.

\subsection{Generating Valid Models}
\label{sec:valid_models}

With the rules for stacking layers together in place, generating valid models is straightforward. An initial layer type has to be specified by the user, which can be fully connected, convolutional or recurrent. Defining the initial layer type effectively defines the type of architectures that can be generated by the algorithm. That is, if the user chooses fully connected as the initial layer, all the generated models will be fully connected. If the user chooses convolutional as initial layer, the algorithm will generate convolutional models only and so on.

Just as the initial layer type has to be user-defined, the final or output layer is also user-defined. In fact, all  the generated models share the same output layer. The output layer is always a fully connected layer. Furthermore, it is generated based on the type of problem to solve, i.e. classification or regression. In the case of classification, the number of neurons is defined by the number of classes in the problem and the softmax function is used as activation function. For regression problems, the number of neurons is one and the activation function used is the linear function.

Having defined the architecture type and the output layer, generating an initial model is an iterative process of stacking new layers that comply with the rules in Table \ref{table:neural_network_building_rules}. A user-defined parameter $\gamma_l$ is used to stop inserting new layers. Every time a new layer is stacked in the model, a random number $\psi \in \left[0,1\right]$ is generated using the following probability distribution
\begin{equation}
\rho_l = 1 - \sqrt{1 - U},
\label{eq:more_layers_prob}
\end{equation}
where $U$ is a uniformly distributed random number. If $\rho_l < \gamma_l$ and if the current layer is compatible with the last layer according to Table \ref{table:neural_network_building_rules}, then no more layers are inserted. Equation (\ref{eq:more_layers_prob}) is used to let the user choose the probability with which more layers are stacked to a neural network model. Thus, if the user wants that a new layer is inserted with $80\%$ probability, the user must choose $\gamma_l = 0.8$. 

With regards to layers that have an activation function, even though in principle any valid activation functions are allowed, for this application we choose to keep the same activations for similar layers across the model since this is usually the common practice. 

\subsection{Selection}
\label{sec:selection}

In order to generate $n$ offsprings, $2n$ parents are required. The parents are chosen using a selection mechanism which takes the population $\mathcal{C}(t)$ at the current generation and returns a list of parents for crossover. For our application, the selection mechanism used is based on the binary tournament selection \cite{Engelbrecht2007,Krishnakumar1989}. A description of the mechanism is given next.

\begin{itemize}
\item Select $m$ parents at random where $m < n$.
\item Compare the selected elements in a pair-wise manner and return the most fit individuals.
\item Repeat the procedure until $2n$ parents are selected.
\end{itemize}

It is important to note in the above procedure that the larger $m$ is, the higher the probability that the best individual in the population is chosen as one of the parents, this is not a desirable behavior, thus we warn the users to keep $m$ small. Also, since our algorithm uses elitism best individual of a current generation remains unchanged in the next generation.

\subsection{Crossover Operator}
\label{sec:crossover}

Since the encoding chosen for this task is rather peculiar, the existing operators are not suitable for it. We design a new crossover operator. In this section, we describe in detail the proposed crossover operator. The operator is based on the two point crossover operator for genetic algorithms \cite{holland1992} in the sense that two points are selected for each parent. The operator is more restrictive as to which pairs of points may be selected in order to ensure the generation of valid architectures. 

The key concept behind our crossover operator is that of  ``compatibility between pairs of points''. Consider two models $S_1$ and $S_2$ that will serve as parents for one offspring. Assume that the offspring will be generated by replacing some layers in $S_1$ from some layers in $S_2$.  $S_1$ is thus the base parent. If we select any two pairs of points $(r_1, r_2)$ from $S_1$ and $(r_3, r_4)$ from $S_4$, it may happen that such pairs of points cannot be interchangeable because layer $r_3$ cannot be placed instead of layer $r_1$ or layer $r_4$ cannot be placed instead of layer $r_2$. Therefore, the selection mechanism must ensure that the interchange points, $(r_1, r_2), (r_3, r_4)$, are compatible. That is to say, layer $r_3$ is compatible with the layer preceding $r_1$ and the layer after $r_2$ is compatible with layer $r_4$. Compatibility is defined in terms of the rules described in Table \ref{table:neural_network_building_rules}. A selection mechanism that guarantees the compatibility between pairs of points is described in Algorithm \ref{algorithm:crossover_method}. This method assume that the offspring will be generated by replacing some layers in $S_1$ from some layers in $S_2$.

\begin{algorithm}[H]
\SetAlgoLined
\KwData{None}
\SetKwInOut{Input}{Input}\SetKwInOut{Output}{Output}

\Input{Two neural network string representations $S_1$ and $S_2$.}
\Output{A neural network string representation $S_3$.}

\BlankLine

Let $S_1$ and $S_2$ be the arrays containing the stacked layers of a neural network model in parents 1 and 2, respectively.
Take two random points $(r_1, r_2)$ from $S_1$ where $r_1 <= r_2 $.\\
\If{$r_1 = r_2 $}{$r_2 = len(S_1-1)$} 
\Else{pass}
Find all the pairs of points $(r_3, r_4)_i$ in $S_2$ that are compatible with $(r_1, r_2)$ where $r_3 < r_4$ and $(r_4 - r_3) - (r_2 - r_1) < \mathcal{A}$.\\
Randomly pick any of pairs $(r_3, r_4)_i$.\\
Replace the layers in $S_1$ between $r_1, r_2$ inclusive with the layers in $S_2$ between $r_3, r_4$ inclusive. Label the new model as $S_3$.\\
Rectify the activation functions of $S_3$ to match the activation functions of $S_1$.

\caption{Crossover method.}
\label{algorithm:crossover_method}
\end{algorithm}

It is possible that the mechanism described in Algorithm \ref{algorithm:crossover_method} requires more than one attempt to find valid interchange points $(r_1, r_2)$ and  $(r_3, r_4)$ for models $S_1$ and $S_2$. Based on our experience with the method and the obtained results,  Algorithm \ref{algorithm:crossover_method} usually requires only one attempt to successfully generate a valid offspring. To prevent the crossover mechanism from getting trapped in an infinite loop, we limit the number of trials to $\gamma_c$ where $\gamma_c=3$ is the default and can be adjusted by the user. Let us illustrate Algorithm \ref{algorithm:crossover_method} with an example. Consider the following models
\begin{align*}
S_1 & = \left[ \left[1, 264, 2, 0, 0, 0, 0, 0 \right], \left[5, 0, 0, 0, 0, 0, 0, 0.65 \right], \right. \\
& \left. \left[1, 464, 2, 0, 0, 0, 0, 0 \right], \left[5, 0, 0, 0, 0, 0, 0, 0.35 \right], \right. \\
& \left. \left[1, 872, 2, 0, 0, 0, 0, 0 \right], \left[1, 10, 3, 0, 0, 0, 0, 0 \right] \right]
\end{align*}

\begin{align*}
S_2 & = \left[ \left[1, 56, 0, 0, 0, 0, 0, 0 \right], \left[5, 0, 0, 0, 0, 0, 0, 0.25 \right], \right. \\
&  \left. \left[1, 360, 0, 0, 0, 0, 0, 0 \right], \left[1, 480, 0, 0, 0, 0, 0, 0 \right] \right. \\
&  \left. \left[1, 88, 0, 0, 0, 0, 0, 0 \right], \left[5, 0, 0, 0, 0, 0, 0, 0.2 \right], \right. \\
&  \left. \left[1, 10, 3, 0, 0, 0, 0, 0 \right] \right]
\end{align*}

Let us take $r_1 = 1$ and $r_2 = 3$, since these points are going to be removed from the model we need to find the compatible layers with $S_1[r_1-1]$ and $S_1[r_2]$ according to the rules described in Table \ref{table:neural_network_building_rules}. Note that if $r_1 = 0$, i.e. the initial layer, only a layer whose layer type is equal to the layer type of $S_1[0]$ is compatible. Thus, for this example, the compatible pairs of points $(r_3, r_4)_i$ are
\begin{align*}
& \left[ (0, 0), (0, 2), (0, 4), (0, 5), (1, 2), (1, 4), (1, 5), \right. \\
& \left. (2, 2), (2, 4), (2, 5), (4, 4), (4, 5), (5, 5) \right]
\end{align*} 
Assume that we pick at random the pair $(2,4)$. The offspring, which we will call $S_3$, looks like
\begin{align*}
S_3 & = \left[ \left[1, 264, 2, 0, 0, 0, 0, 0 \right], \left[1, 360, 2, 0, 0, 0, 0, 0 \right], \right. \\
& \left. \left[1, 480, 2, 0, 0, 0, 0, 0 \right] , \left[1, 88, 2, 0, 0, 0, 0, 0 \right], \right. \\
& \left. \left[1, 872, 2, 0, 0, 0, 0, 0 \right], \left[1, 10, 3, 0, 0, 0, 0, 0 \right] \right]
\end{align*}
which is a valid model. The reader is encouraged to check the actual neural network representations for each of the models in Appendix \ref{appendix:generated_models}. Notice that all the activation functions of the same layer types are changed to match the activation functions of the first parent $S_1$. We call this process ``activation function rectification''. It basically implies changing all the activation functions of the layers that share the same layer type between $S_1$ and $S_3$ to the activation functions used in $S_1$. 

Finally, one important feature of this crossover operator is that it has the ability to generate neural network models of different sizes, i.e. it can shrink or increase the size of the base parent. This behavior mimics that of machine learning practitioners, which will often start with a base model and iteratively shrink or increase the size of the base model in order to find the one that has the best inference performance.

\subsection{Mutation Operator}
\label{sec:mutation}

The mutation operator is used to induce small changes to some of the models generated through the crossover mechanism. In the context of evolutionary computation, these subtle changes tend to improve the exploration properties of the current population, i.e. to keep genetic diversity, by injecting random noise to the current solutions.  Although mutation is not needed in the micro-GA according to \cite{Krishnakumar1989}, we believe some sort of mutation is beneficial for our application to get more diverse models which could potentially lead to better inference abilities. Nevertheless, our mutation approach will be less aggressive in order to mitigate its effect. In the following, we discuss the core ideas of our mutation mechanism.

As stated above, our mutation approach is less aggressive than common mutation operators \cite{Engelbrecht2007}. Our design follows two main reasons: First, the fact that usually micro genetic algorithms don't make use of the mutation operator since the crossover operator has already induced significant genetic diversity in the population, thus we want to minimize its impact. The second reason is related to the way neural networks are usually built by human experts. Usually, experts try a number of models and then make subtle changes to each of them in order to improve their inference ability. Such changes usually involve adjusting the parameters in a layer, adding or removing a layer, adding regularization or changing the activation functions. Based on these reasons, our mutation process randomly chooses one layer of the model, using a probability $\rho_m$,  and then proceeds to make one of the following changes to it:
\begin{itemize}
\item Change a parameter of the layer chosen for a value complying the values stated in Table \ref{table:neural_network_array_details}.
\item Change the activation function of the layer. This would involve rectifying the entire model as described in Section \ref{sec:crossover}.
\item Add a dropout layer if the chosen layer is compatible.
\end{itemize} 

These operations altogether provide a rich set of possibilities for performing efficient mutation while still generating valid models after mutation is performed. Furthermore, the original model is barely changed.

\subsection{Nominal Convergence}
\label{sec:convergence}

Nominal convergence is one of the criteria used for early stopping of the proposed evolutionary algorithm. Some literature defines the convergence in terms of the fitness of the individuals \cite{Engelbrecht2007}, while in \cite{Krishnakumar1989} the convergence is defined in terms of the genotype or phenotype of the individuals. Although convergence based on the actual fitness of the individuals may be easier to assess given that the fitness is already calculated, we believe that an assessment of convergence based on the actual genotype of the individuals suits the current needs better.

Since neural networks are stochastic in nature, we expect some variations in the fitness of the individuals at every different run. Furthermore since we are running the training process for only few epochs, the final performance of the networks can be quite different and would not be a reliable indicator of convergence. Instead, to assess the convergence we, look at the genotype of the actual neural network architecture and compute the layer-wise distance between the different individuals in population $\mathcal{C}$.

Let $S_1$ and $S_2$ be the genotypes representing any two different models such that  where $len(S_2) >= len(S_1)$. Let $S_1[j]$ be the vector representation of the $j^{th}$ layer of model $S_1$. The method for computing the distance $d(S_1, S_2)$ between any two models $S_1$ and $S_2$ is defined in Algorithm \ref{algorithm:difference_between_networks}.

\begin{algorithm}[H]
\SetAlgoLined
\KwData{None}
\SetKwInOut{Input}{Input}\SetKwInOut{Output}{Output}

\Input{Two neural network string representations $S_1$ and $S_2$.}
\Output{Distance between models $S_1$ and $S_2$}

\BlankLine

Let $\mathbf{d} \in \mathbb{R}$ be the distance between $S_1$ and $S_2$. Make $\mathbf{d} = 0$.\\
\For{Each layer $i$ in $S_1$ except last layer}{$d = d + norm_2(S_2[i]-S_1[i])$.}
\For{Each remaining layer $i$ in $S_2$ except last layer}{$d = d + norm_2(S_2[i])$.}
Return $\mathbf{d}$.
\caption{Layer-wise distance $d(S_1, S_2)$ between model genotypes.}
\label{algorithm:difference_between_networks}
\end{algorithm}

This method is computationally inexpensive since the size of the population is small. Furthermore, it helps accurately establish the similarity between two neural network models. Given two neural network models $S_1$ and $S_2$, if $d(S_1, S_2) = 0$, then $S_1 = S_2$. We say that AMS (Algorithm \ref{algorithm:nn_ea}) has reached a nominal convergence if at least $m_c$ pairs of models have $d(S_1, S_2) \leq d_t$, where both $m_c$ and $d_t$ are user defined parameters.

\subsection{Implementation}
\label{sec:implementation}

AMS is implemented in about 700 lines of code in Python. The code can be found in \url{https://github.com/dlaredo/automatic_model_selection} \cite{Laredo2019}. We took an object oriented programming approach for its implementation. The models $S_i$ generated by the algorithm are fetched to Keras \cite{keras2015} and then evaluated. The models can be fetched and evaluated in any other framework such as TensorFlow or Pytorch and even other programming languages including C++.

In order to boost performance of AMS, we make use of Ray \cite{Moritz2017} which is a distributed computing framework tailored for AI applications. In order to distribute workloads in Ray, developers only have to define Remote Functions by making use of Python annotations. Ray will then distribute these Remote Functions across the different nodes in the cluster. There are three different types of nodes in Ray: Drivers, Workers and Actors. A Driver is the process executing the user program, a Worker is a stateless process that executes remote functions invoked by a driver or another worker, and finally, an Actor is a stateful process that executes, when invoked, the methods it exposes.

For the implementation, we code the individual fetching to Keras and its fitness evaluation as Ray Remote Functions, i.e. Workers, while the rest of the algorithm is implemented within the Driver. Partial training of each neural network within the current population can therefore be performed in a distributed way, leading to a highly increased performance of the algorithm. Furthermore, since the only messages being sent over the cluster are arrays of the neural network representation $S_i$ and the performance $p$ of the neural network model, there is little chance that the interchange of data causes a bottleneck or increases latency in the system. Nevertheless, each node in the cluster has to maintain a local copy of the dataset.

\section{Evaluation}
\label{sec:evaluation}

We evaluate AMS with two different datasets, each of which represents a different type of inference problem. We also compare our results with the state-of-the-art neural network models for each problem. For the experiments in this section, each model generated by AMS is trained using the following parameters.

\begin{table}[H]
\begin{center}
\scalebox{1}{
\begin{tabular}{| c | c | c | c | c | c |}
\hline
Dataset & Epochs & Learning Rate & Optimizer & Loss Function & Metrics \\
\hline
{MNIST} & {5} & {0.001} & {Adam} & Categorical & Categorical\\
  &     &  &  & cross-entropy & accuracy\\
CMAPSS & 5 & 0.01 & Adam & MSE & MSE\\
\hline
\end{tabular}
}
\end{center}
\caption{Training parameters for each of the used datasets.}
\label{table:training_params}
\end{table}

For the CMAPSS dataset, we use a larger learning rate since we intend to evaluate the model using very few epochs for this complex problem. In order to get a clear idea of which individuals within the population may be promising, we make the learning process more aggressive during the first iterations of the algorithm. 

All of the experiments were run using the Keras/Tensorflow GPU framework. A desktop PC with a Core i7-8700k processor and an NVIDIA GeForce 1080Ti GPU. Ray was not used for the results presented here.

\subsection{MNIST Dataset - A Classification Problem}

We first test our algorithm on the MNIST dataset \cite{Lecun1989}. The MNIST dataset of handwritten digits is one of the most commonly used datasets for evaluating the performance of neural networks. It has a training set of 60,000 examples and a test set of 10,000 examples. The digits have been size-normalized and centered in a fixed-size image, while the size of each image is of 28x28 pixels. As a part of the data pre-processing, we normalize all the pixels in each image to be in the range of $[0,1]$ and unroll the 28x28 image into a vector with 784 components.

We use MNIST dataset as a baseline for measuring the performance of the proposed approach. Furthermore, we use MNIST to analyze each one of the major components of AMS. Given the popularity of MNIST, several neural networks with varying degrees of accuracy have been proposed in the literature. Therefore, it is easy to find good models to compare with. 

We start by running AMS to find a suitable fully connected model for classification of the MNIST dataset. The details for the parameters used in this test are described in Table \ref{table:MNIST_params}. Each of the experiments carried out by AMS takes about 4 minutes in our test computer. 

\begin{table}[H]
\begin{center}
\scalebox{0.8}{%
\begin{tabular}{| c | c |}
\hline
Parameter & AMS Value \\
\hline
Problem Type & 1 \\
Architecture Type & FullyConnected \\
Input Shape & $(784, M)$  \\
Output Shape & $(10, M)$ \\
Cross Validation Ratio & $\gamma_v = 0.2$ \\
Mutation Probability & $\rho_m = 0.4$ \\
More Layers Probability & $\gamma_l = 0.4$ \\
Network Size Scaling Factor & $\alpha = 0.5$ \\
Population Size & $n = 10$ \\
Tournament Size & $n_t = 4$ \\
Max Similar Models & $\gamma_c = 3$ \\
Training epochs & $\gamma_t = 5$\\
Max generations & $\gamma_g = 10$ \\
Total Experiments & $\gamma_r = 5$ \\
\hline
\end{tabular}
}
\end{center}
\caption{AMS Parameters for MNIST dataset.}
\label{table:MNIST_params}
\end{table}

We first take a look at the generated initial population. For the sake of space, we will only discuss the sizes of the models. Furthermore, we make a small change in our notation for describing neural network models. For the remainder of this section, we denote a layer of a neural network as $(n_e, a_f)$ where $n_e$ denotes the number of neurons for fully connected layer or the dropout rate for a dropout layer, and $a_f$ denotes the activation function of the fully connected layer. The initial five generated individuals are presented below. Fitness, accuracy and raw size of the models in the initial population are presented in Table \ref{table:ams_mnist_initial}.

\begin{align*}
S_1 & = \left[ (64, 0), (0.4), (10, 3) \right] \\
S_2 & = \left[ (760, 2), (0.5), (608, 2), (0.65), (10, 3) \right] \\
S_3 & = \left[ (864, 0), (0.15), (536, 0), (928, 0), (10, 3) \right] \\
S_4 & = \left[ (40,0), (0.45), (912, 0), (10, 3) \right] \\
S_5 & = \left[ (968, 1), (976, 1), (32, 1), (0.15), (808, 1), \right.\\
& \left. (10, 3), (832, 2) \right] \\
\end{align*}

\vspace{-2em}

\begin{table}[H]
\begin{center}
\begin{tabular}{| c | c | c | c |}
\hline
Model & Score (Accuracy) & Trainable Parameters & Fitness\\
\hline
$S_1$ & 91.7\% & 50890 & 2.7688\\
$S_2$ & 98.7\% & 1065378 & 3.0812\\
$S_3$ & 94.6\% & 1649506 & 3.3791\\
$S_4$ & 92.1\% & 77922 & 2.8427\\
$S_5$ & 96.6\% & 1771642 & 3.2867\\
\hline
\end{tabular}
\end{center}
\caption{Scores for the initial population found by AMS for MNIST. $\alpha = 0.5$}
\label{table:ams_mnist_initial}
\end{table} 

Observe that the sizes of the models in the initial population are diverse with some models having as few as 1 hidden layer and some having more than 5 hidden layers. The number of layers of the models in the initial population is defined by the parameter $n_r$. We set this value to be small on purpose, since MNIST is a dataset that is easy even for simple neural network models. Also note that the initial population has models with different activation functions. The Sigmoid, Tanh and ReLU are all used by some models. This is beneficial to the search process as some activation functions may yield better results than others.

Finally, we note that some models in the initial population already yield decent accuracy (about $90\%$). They also have a large number of trainable parameters. It is decided that, in the case of MNIST dataset, the task for AMS is to find a model with an accuracy higher than $95\%$ and a small number of trainable parameters.

For a value of the network size scaling factor $\alpha = 0.5$ in Equation (\ref{eq:fitness_function_scalar}), after 5 experiments and 10 generations for each experiment, AMS converges to the following five models. As a side note, for all of the experiments in this section we denote the best model found at experiment $i \in \left\lbrace 1, \cdots, n \right\rbrace$ as $S_i^*$ and the best model out of $n$ experiments for a given $\alpha$ as $S_i^+$. The fitness, accuracy and raw size of the models are presented in Table \ref{table:ams_mnist_1}.

\begin{align*}
S^*_1 & = \left[ (56, 2), (10, 3) \right] \\
S^*_2 & = \left[ (168, 2), (10, 3) \right] \\
S^*_3 & = \left[ (40, 2), (0.2), (10, 3) \right] \\
S^+_4& = \left[ (48, 2), (48, 2),  (10, 3) \right] \\
S^*_5 & = \left[ (312, 2), (10, 3) \right] \\
\end{align*}

\vspace{-2em}

\begin{table}[H]
\begin{center}
\begin{tabular}{| c | c | c | c |}
\hline
Model & Score (Accuracy) & Trainable Parameters & Fitness\\
\hline
$S^*_1$ & 93.9\% & 44530 & 2.6287\\
$S^*_2$ & 95.8\% & 133570 & 2.7736\\
$S^*_3$ & 93.4\% & 31810 & 2.5826\\
$S^+_4$ & 94.7\% & 40522 & 2.5693\\
$S^*_5$ & 95.0\% & 248050 & 2.9431\\
\hline
\end{tabular}
\end{center}
\caption{Scores for the best models found by AMS for MNIST. $\alpha = 0.5$}
\label{table:ams_mnist_1}
\end{table}

Table \ref{table:ams_mnist_1} shows a clear preference for small models. Furthermore, there seems to be a preference for ReLU activation functions. It can also be observed that for $\alpha = 0.5$, a good balance between size of the network and its performance is obtained. In the following, we perform tests with $\alpha = 0.3$ and $\alpha = 0.7$ to further analyze the behavior of the algorithm with respect to the preference of the network size scaling factor. A smaller $\alpha$ value will prioritize a better performing network while a larger value of $\alpha$ instructs AMS to focus on light-weight models.

The best models for each experiment with $\alpha = 0.3$ are listed below. The fitness and raw size of the models are described in Table \ref{table:ams_mnist_2}. 

\begin{align*}
S^*_1 & = \left[ (32, 0), (10, 3) \right] \\
S^*_2 & = \left[ (32, 1), (32, 1), (10, 3) \right] \\
S^+_3 & = \left[ (64, 1), (64, 1), (64, 1), (64, 1), (10, 3) \right] \\
S^*_4 & = \left[ (56, 1),  (10, 3) \right] \\
S^*_5 & = \left[ (40, 1), (10, 3) \right] \\
\end{align*}

\vspace{-2em}

\begin{table}[H]
\begin{center}
\begin{tabular}{| c | c | c | c |}
\hline
Model & Score (Accuracy) & Trainable Parameters & Fitness\\
\hline
$S^*_1$ & 91.5\% & 25450 & 1.9126\\
$S^*_2$ & 95.2\% & 26506 & 1.6677\\
$S^+_3$ & 97.2\% & 63370 & 1.6358\\
$S^*_4$ & 94.7\% & 44530 & 1.7640\\
$S^*_5$ & 94.4\% & 31810 & 1.7412\\
\hline
\end{tabular}
\end{center}
\caption{Scores for the best models found by AMS for MNIST, $\alpha = 0.3$.}
\label{table:ams_mnist_2}
\end{table}

The models presented in Table \ref{table:ams_mnist_2} exhibit, in general, better performance than those depicted in Table \ref{table:ams_mnist_1}. This is due to the fact that $\alpha = 0.3$ prioritizes the model accuracy over model size. Surprisingly, the sizes of the models obtained when $\alpha = 0.3$ are on the same order of magnitude as those obtained when $\alpha = 0.5$. A discussion on this behavior is provided later in this section.

We repeat the experiment with $\alpha = 0.7$. The obtained models are listed below and their fitness, raw size and accuracy are shown in Table \ref{table:ams_mnist_3}.

\begin{align*}
S^*_1 & = \left[ (24, 2), (10, 3) \right] \\
S^*_2 & = \left[ (104, 2), (0.2), (10, 3) \right] \\
S^+_3 & = \left[ (16, 2), (24, 2), (10, 3) \right] \\
S^*_4 & = \left[ (56, 2), (10, 3) \right] \\
S^*_5 & = \left[ (208, 2),  (10, 3) \right] \\
\end{align*}

\vspace{-2em}

\begin{table}[H]
\begin{center}
\begin{tabular}{| c | c | c | c |}
\hline
Model & Score (Accuracy) & Trainable Parameters & Fitness\\
\hline
$S^*_1$ & 92.2\% & 19090 & 3.2284\\
$S^*_2$ & 95.2\% & 82690 & 3.5856\\
$S^+_3$ & 92.1\% & 13218 & 3.1158\\
$S^*_4$ & 94.5\% & 44530 & 3.4200\\
$S^*_5$ & 95.8\% & 165370 & 3.7762\\
\hline
\end{tabular}
\end{center}
\caption{Scores for the best models found by AMS for MNIST, $\alpha = 0.7$.}
\label{table:ams_mnist_3}
\end{table}

The results in Table \ref{table:ams_mnist_3} show that when the number of trainable parameters has a large impact on the overall fitness of the individuals, and the algorithm tends to prefer smaller networks, which is specially useful for cases where the computational power is limited, such as embedded systems. This brings a dilemma, namely that neural networks that exhibit a lower performance as compared to larger neural networks may be preferred. Nevertheless, this trade-off can be controlled by the user by varying the $\alpha$ parameter. 

As a side note, we point out that the difference in the fitness exhibited among the models of the three different experiments is due to the fact that the size of the neural network is scaled as described in Section \ref{sec:fitness_function}. Thus, there is no fair way to compare the fitness of the models shown in Tables \ref{table:ams_mnist_1}, \ref{table:ams_mnist_2} and \ref{table:ams_mnist_3} against each other. We are clearly dealing with a multi-objective optimization problem with conflicting objective functions.

\begin{figure}[H]
\centering
\includegraphics[scale=0.7]{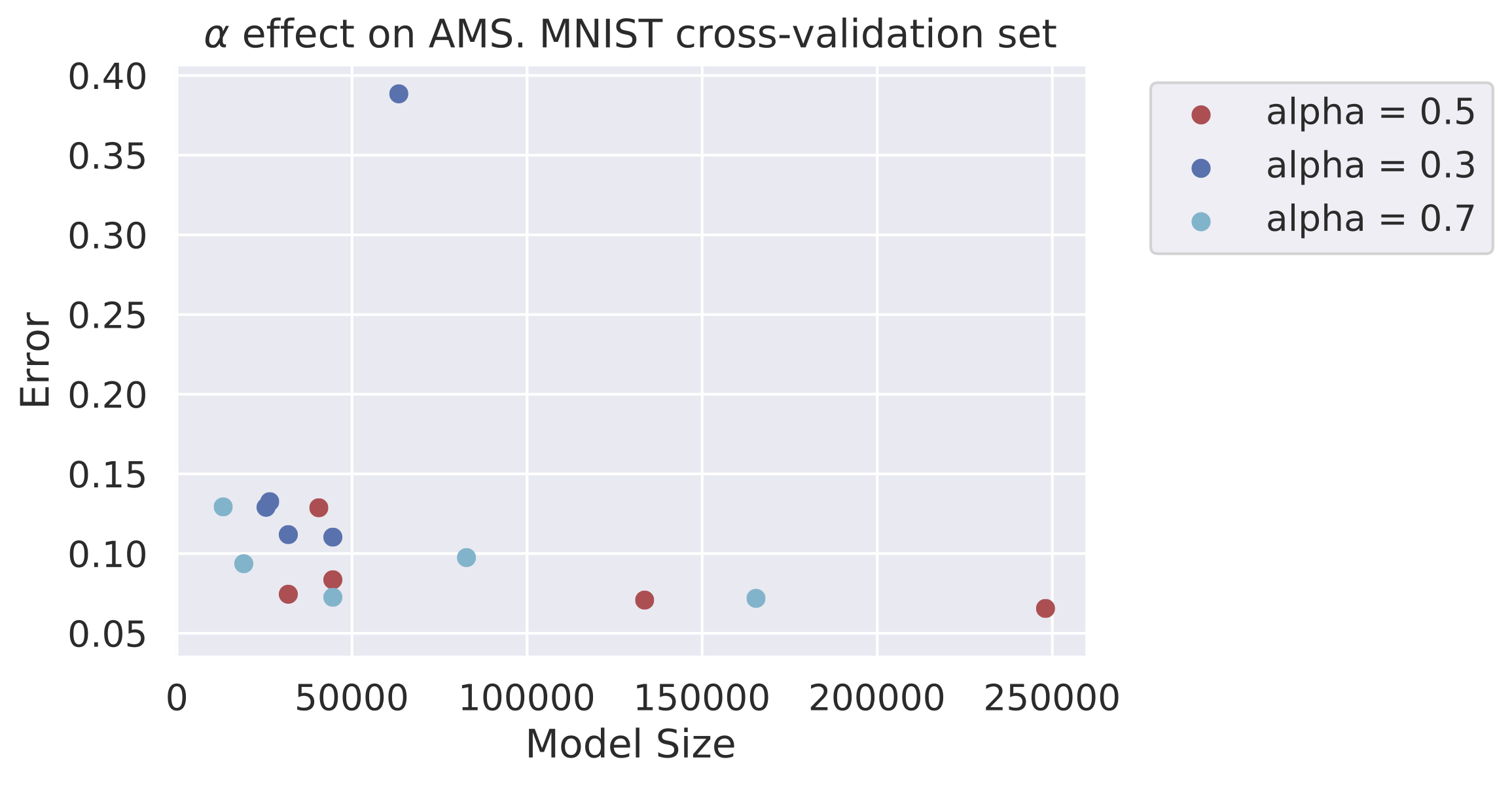}
\caption{Cluster formed by the found models for different $\alpha$ values for MNIST cross-validation set.}
\label{fig:alpha_mnist_cluster_cvset}
\end{figure}

Figure \ref{fig:alpha_mnist_cluster_cvset} shows the results obtained by all of the models for $\alpha \in \left\lbrace 0.3, 0.5, 0.7  \right\rbrace$, which are trained for 100 epochs using 5-fold cross-validation to assess their accuracy. It is observed that the models cluster around a model size less than 50000 and an error less than 0.15. As expected, the models obtained with $\alpha = 0.7$ yield, in general, the smallest sizes. Outliers are mostly due to the fact that for such experiments the algorithm is unable to find a smaller model, which is likely due to a bad initial population for such experiments. It could also be attributable to the scaling done to the network size as seen in Equation  (\ref{eq:fitness_function_scaled}). Since AMS uses a logarithmic scale to measure the size of the networks, a few thousands of weights are unlikely to make a big difference in the fitness of a model. 

Finally, we compare the best models for each value of $\alpha$ against each other. A 10-fold cross-validation process, with a training of 50 epochs per fold, is carried out for each one of the best models in order to obtain the mean measure of accuracy for each model. The three models are feed the same training data and are validated using the same folds for the cross-validation data. We also measure the accuracy of each models by using a test set that is never used during the training or hyper-parameter tuning processes of the models. The accuracy averages and size of the networks are summarized in Table \ref{table:avg_accuracies_mnist}. 

\begin{table}[H]
\begin{center}
\begin{tabular}{| c | c | c | c |}
\hline
Model & 5-fold Avg. Score & Test Accuracy & Network size\\
\hline
$S^+_{0.3}$ & 97.3\% & 97.4\% & 63370\\
$S^+_{0.5}$ & 96.8\% & 97.0\% & 40522\\
$S^+_{0.7}$ & 95.0\% & 95.4\% & 13218\\
\hline
\end{tabular}
\end{center}
\caption{Accuracy obtained by each of the top 3 models for MNIST dataset.}
\label{table:avg_accuracies_mnist}
\end{table}

As expected, the model obtained for $\alpha = 0.7$ yields the smallest neural network, about 4 times smaller than the model obtained when $\alpha = 0.3$. Nevertheless, its accuracy is the worst of the three models, though only by a small margin. On the opposite, the resulting model for $\alpha = 0.3$ gets the best performance in terms of accuracy, but also attains the largest neural network model of the three. Finally, when $\alpha = 0.5$ AMS yields a model with a good balance between the model size and performance. It is important to highlight that to obtain the models presented in Table \ref{table:avg_accuracies_mnist}, at most 50 different models are tried. Nevertheless, since each model is trained for only 5 epochs, the total time spent by the algorithm to find the models in Table \ref{table:avg_accuracies_mnist} is less than 4 minutes in our test computer. 

\begin{figure}[H]
\centering
\subfigure[Obtained models on cv set]{
\includegraphics[scale=0.4]{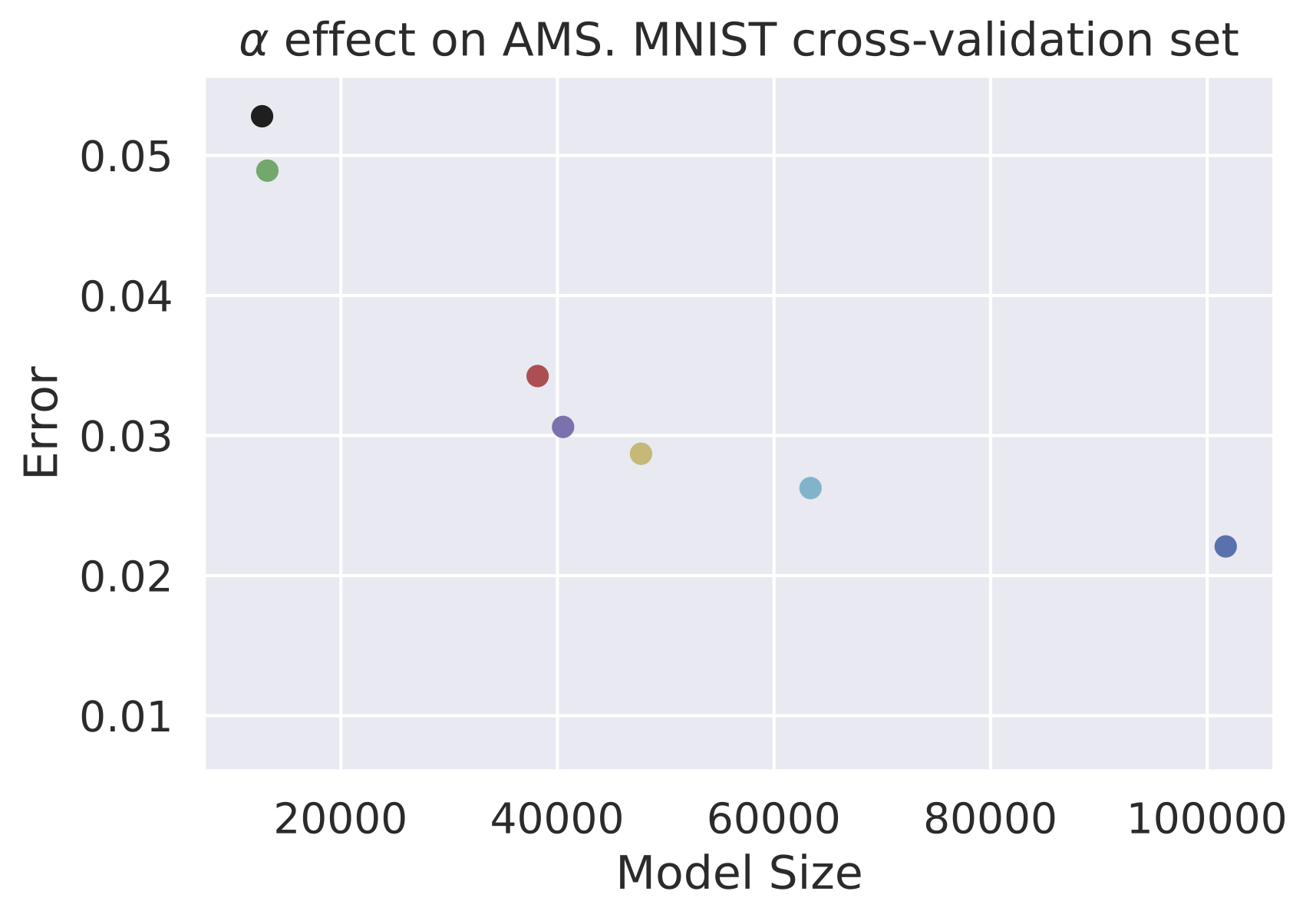}
\label{fig:alpha_mnist_cvset}
}
\subfigure[Obtained models on test set]{
\includegraphics[scale=0.4]{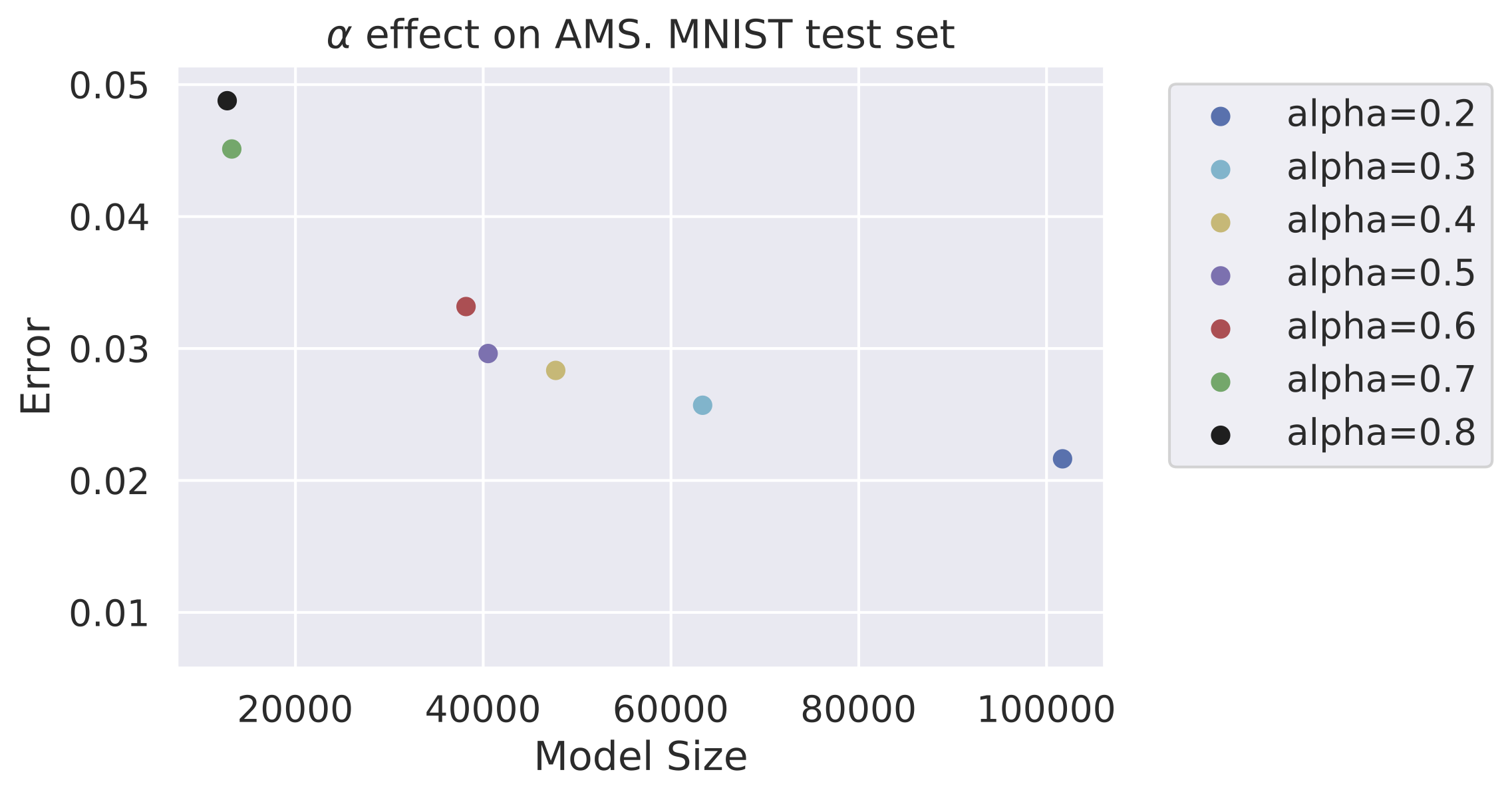}
\label{fig:alpha_mnist_testset}
}
\caption{Influence of $\alpha$ on the model's size and error on MNIST dataset}
\label{fig:alpha_mnist}
\end{figure}

Figure \ref{fig:alpha_mnist} plots the size of the model vs. the error. As can be observed, the resulting models for different $\alpha$ values form a so called Pareto front \cite{Nocedal06}, i.e., none of the resulting models is better than the others in terms of the size and error. The trade-off between the size and performance of the model can be clearly seen in the figure. Although this is a nice property exhibited by the MNIST dataset, this need not always hold since it is known that the performance of a neural networks does not monotonically increase with the size of the model. 

Table \ref{table:MNIST_results_top} shows three of the top hand-crafted models along with their obtained accuracy for the MNIST dataset, it can be observed that the accuracy obtained by AMS is close to that obtained by the fine tuned models. 

\begin{table}[H]
\begin{center}
\begin{tabular}{| c | c | c |}
\hline
Method & Test accuracy & Size\\
\hline
3 Layer NN, 300+100 hidden units \cite{Lecun1989} &  96.95\% & 266610\\
2 Layer NN, 800 hidden units \cite{Simard2003} & 98.4\% & 636010\\
6-layer NN (elastic distortions) \cite{Ciresan2010}  & 99.65\% & 11972510\\
\hline
\end{tabular}
\end{center}
\caption{Top results for MNIST dataset.}
\label{table:MNIST_results_top}
\end{table}

By comparing the models obtained by AMS in Table \ref{table:avg_accuracies_mnist} against the models in Table \ref{table:MNIST_results_top} we can observe that, even though AMS models don't attain the highest accuracy, they exhibit good inference capabilities with a much lesser number of trainable parameters, at least one order of magnitude smaller. This shows that AMS models have a good balance between the inference power of the model and its size. Furthermore, the score-size trade-off can be controlled by means of the parameter $\alpha$, where a value closer to $\alpha=0$ makes AMS prefer networks with higher accuracy and a value closer to $\alpha = 1$ makes AMS prefer networks with smaller sizes.

\subsection{CMAPSS Dataset - A Regression Problem}

Here we analyze the performance of AMS when dealing with regression problems. For testing regression, we use the C-MAPSS dataset \cite{Saxena2008a}. The C-MAPSS dataset contains the data produced using a model based simulation program developed by NASA. The dataset is further divided into 4 subsets composed of multi-variate temporal data obtained from 21 sensors, nevertheless for our test we will only make use of the first subset of data. Training and separate test sets are provided. The training set includes run-to-failure sensor records of multiple aero-engines collected under different operational conditions and fault modes.

The data is arranged in an $M\times26$ matrix where $M$ is the number of data points in each subset. The first two variables represent the engine and cycle numbers, respectively. The following three variables are operational settings which correspond to the conditions in Table \ref{table:dataset_cmaps_params} and have a substantial effect on the engine performance. The remaining variables represent the 21 sensor readings that contain the information about the engine degradation over time.

\begin{table}[!htb]
\begin{center}
\caption{CMAPSS dataset details.}
\label{table:dataset_cmaps_params}
\vspace{12pt}
\scalebox{1}{
\begin{tabular}{| c | c | c | c |}
\hline
Train trajectories & Test trajectories & Operating conditions & Fault modes \\
\hline
100 & 100 & 1 & 1 \\
\hline
\end{tabular}
}
\end{center}
\end{table}

Each trajectory within the training and test sets represents the life cycles of the engine. Each engine is simulated with different initial healthy conditions, i.e. no initial faults. For each trajectory of an engine, the last data entry corresponds to the cycle at which the engine is found faulty. On the other hand, when the trajectories of the test sets terminate at some point prior to failure, the need to predict the remaining useful life (RUL) arises. The aim of the MLP model is to predict the RUL of each engine in the test set via regression. The actual RUL values of test trajectories are also included in the dataset for verification. Further discussions of the dataset and details on how the data is generated can be found in \cite{Saxena2008}.

For this test, we follow the data pre-processing described in \cite{Laredo2018}. Only 14 out of the total 21 sensors are used as the input data. Furthermore, we also use a strided time-window with window size of 24, a stride of 1 and early RUL of 129, to form the feature vectors for the MLP. Further details of the time-window approach can be found in \cite{Laredo2018,Li2018}.

We run AMS to find a suitable MLP for regression using the CMAPSS dataset. The parameters used by the algorithm are described in Table \ref{table:CMAPSS_params}.

\begin{table}[H]
\begin{center}
\scalebox{0.8}{%
\begin{tabular}{| c | c |}
\hline
Parameter & AMS Value \\
\hline
Problem Type & 1 \\
Architecture Type & FullyConnected \\
Input Shape & $(336, M)$  \\
Output Shape & $(1, M)$ \\
Cross Validation Ratio & $\gamma_v = 0.2$ \\
Mutation Probability & $\rho_m = 0.4$ \\
More Layers Probability & $\gamma_l = 0.7$ \\
Network Size Scaling Factor & $\alpha = 0.8$ \\
Population Size & $n = 10$ \\
Tournament Size & $n_t = 4$ \\
Max Similar Models & $\gamma_c = 3$ \\
Training epochs & $\gamma_t = 20$\\
Max generations & $\gamma_g = 10$ \\
Total Experiments & $\gamma_r = 5$ \\
\hline
\end{tabular}
}
\end{center}
\caption{Parameters for the CMAPSS dataset.}
\label{table:CMAPSS_params}
\end{table}

We perform experiments for $\alpha \in \left\lbrace 0.3, \cdots, 0.7 \right\rbrace$ with an increment $\Delta_\alpha = 0.1$. For the sake of space, we only discuss the results obtained when $\alpha \in \left\lbrace 0.4, 0.5, 0.6 \right\rbrace$, which are the $\alpha$ values giving the best results. The best models out of the five experiments obtained for each of the $\alpha$ values by AMS are listed in Table \ref{table:avg_rmse_cmapss} along with their RMSE scores and sizes. As with the MNIST dataset, AMS partially evaluates at most 50 different models for each $\alpha$ value. The experiments for each $\alpha$ take about 2 minutes in our test computer.

\begin{align*}
S^+_{0.4} & = \left[ (104, 1), (824, 1), (1, 4) \right] \\
S^+_{0.5} & = \left[ (264, 2), (1, 4) \right] \\
S^+_{0.6} & = \left[ (80, 1), (80, 1), (1, 4) \right] \\
\end{align*}

\vspace{-2em}

\begin{table}[H]
\begin{center}
\begin{tabular}{| c | c | c | c |}
\hline
Model & 5-fold Avg. Score & Test RMSE & Network Size\\
\hline
$S^+_{0.4}$ & 14.87 & 14.99 & 122393\\
$S^+_{0.5}$ & 15.27 & 15.90 & 89233\\
$S^+_{0.6}$ & 14.78 & 15.74 & 33521\\
\hline
\end{tabular}
\end{center}
\caption{RMSE of the top 3 models for the CMAPSS dataset.}
\label{table:avg_rmse_cmapss}
\end{table}

The results presented in Table \ref{table:avg_rmse_cmapss} further demonstrate the impact of the $\alpha$ parameter. As can be observed, the size of the networks grows as $\alpha$ is smaller. It can also be observed that the results obtained by the three proposed models in the cross-validation sets are very close to each other. Again, small networks are usually preferred.
Table \ref{table:CMAPSS_results_top} presents some of the top results obtained by the hand-crafted MLPs for the CMAPSS dataset. The performance of the models is measured in terms of the root mean squared error (RMSE) between the real and predicted RUL values. 

\begin{table}[H]
\begin{center}
\begin{tabular}{| c | c | c |}
\hline
Method & Test RMSE & Network Size \\
\hline
Time Window MLP \cite{Lim2016} & 15.16 & 6041\\
Time Window MLP with EA \cite{Laredo2018} & 14.39 & 7161\\
Deep MLP Ensemble \cite{Zhang2016} & 15.04 &  n/a \\
\hline
\end{tabular}
\end{center}
\caption{Top results for the CMAPSS dataset.}
\label{table:CMAPSS_results_top}
\end{table}

The obtained models are also competitive when compared against some of the latest MLPs designed for the CMAPSS dataset as shown in Table \ref{table:CMAPSS_results_top}. We compare the score obtained in the test set for all the models. It is shown that one of the three models obtained by AMS produces a better score than the two of the three compared models, i.e. Time Window MLP and Deep MLP ensemble. Although the sizes of the neural networks obtained by AMS are one order of magnitude bigger than the hand-crafted models, we can observe that AMS delivered compact models (having few layers with few neurons in each layer).

\begin{figure}[H]
\centering
\subfigure[Obtained models on cv set]{
\includegraphics[scale=0.4]{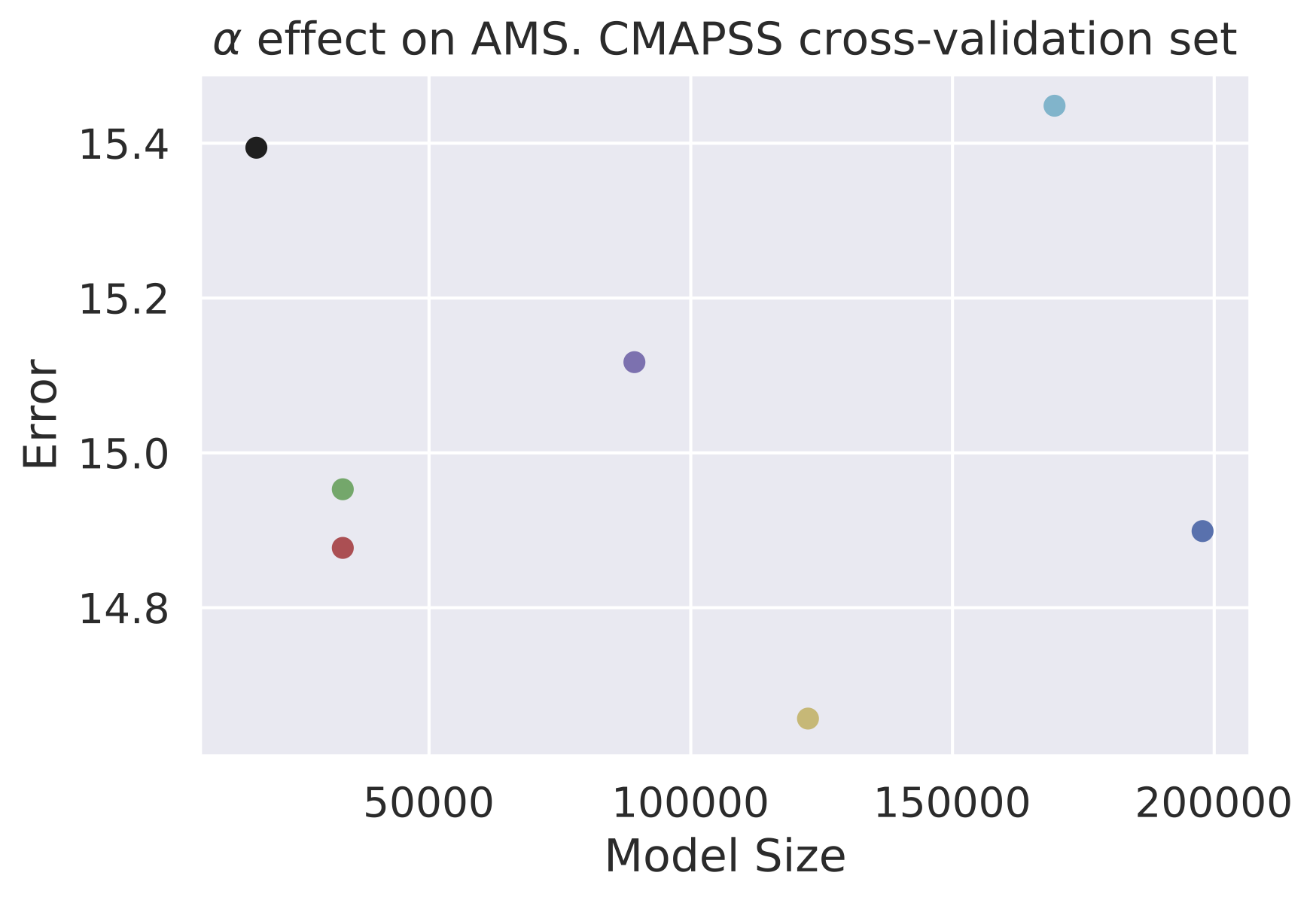}
\label{fig:alpha_cmapss_cvset}
}
\subfigure[Obtained models on test set]{
\includegraphics[scale=0.4]{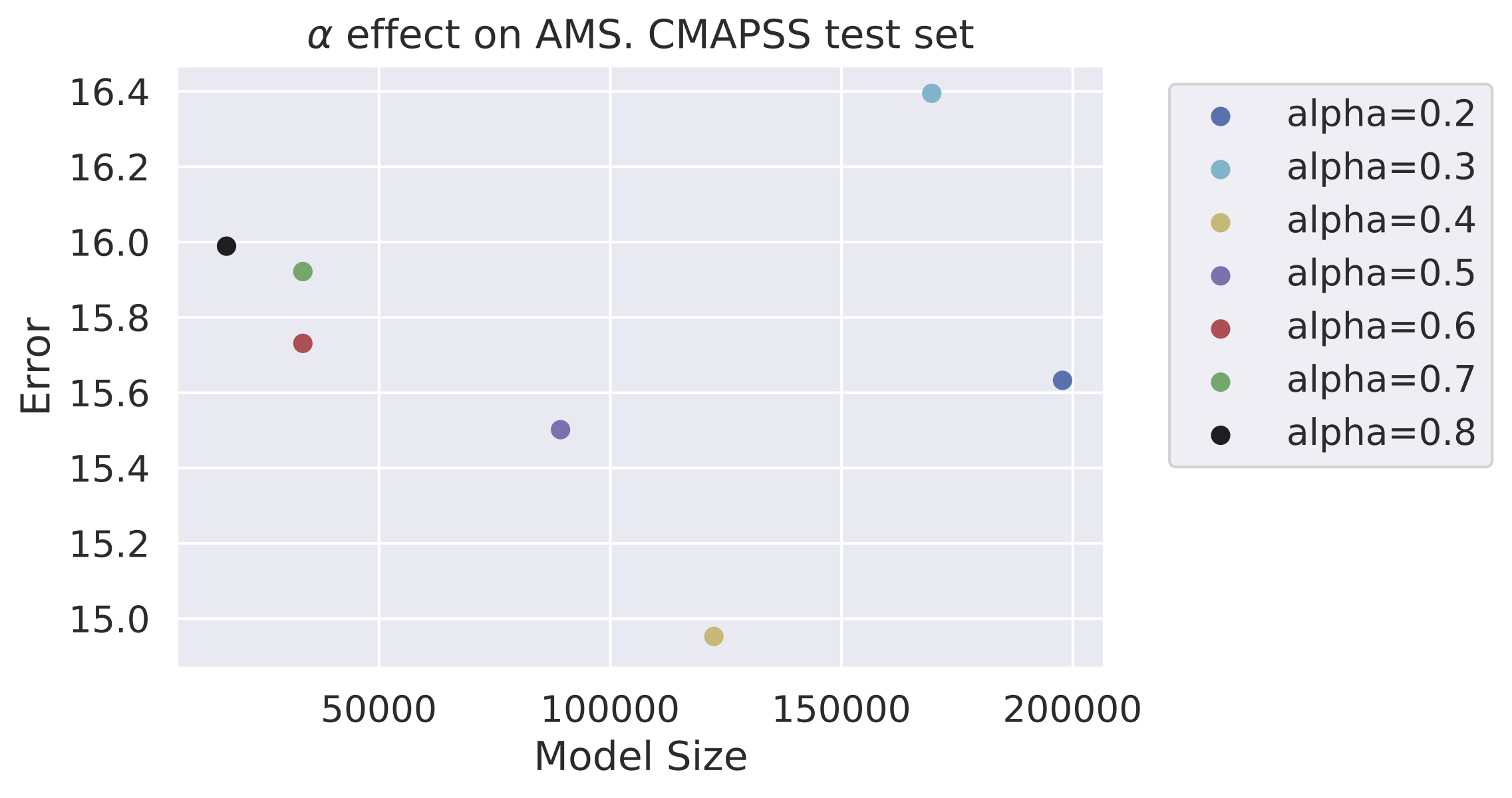}
\label{fig:alpha_cmapss_testset}
}
\caption{Influence of $\alpha$ on the model's size and error on CMAPSS dataset}
\label{fig:alpha_cmapss}
\end{figure}

Finally, Figure \ref{fig:alpha_cmapss} shows that all of the obtained models have a small model size with few hundred thousand parameters while all of them deliver a very good performance for this dataset. See references \cite{Laredo2018} and \cite{Li2018} for other models and their scores. Once again, it can be observed that the model size decreases with larger $\alpha$ values. One important observation is that, for the CMAPSS dataset, there does not seem to be a correlation between the model size and its performance. Indeed, the model size is just one simple indicator of a networks architecture. Trying to characterize a network only by its size may leave out valuable information about it. The information could be used in the search of new individuals with better traits for the dataset. We leave out further analysis of this behavior for future work.

\section{Conclusions and Future Work}
\label{sec:conclusions}

In this paper, we have presented AMS, a new evolutionary algorithm for efficiently finding suitable neural network models for classification and regression problems. Making use of efficient mutation and crossover operators, AMS is able to generate valid and efficient neural networks, in terms of both the size of the network and its performance. Furthermore, AMS design is highly parallelizable and distributable. With the use of frameworks such as Ray \cite{Moritz2017} or Spark \cite{Zaharia2010}, the performance of the algorithm can be greatly boosted.

By allowing the user to control the trade-off between the total size of the network and its performance, AMS is capable of finding small neural networks for applications with limited memory, mobiles or embedded systems that may have constraints on the size of the models or prioritizing the model performance. Although not every model found by AMS is a Pareto point, the found models yield a good balance between the performance and size.

Furthermore, AMS is computationally efficient since it only needs to evaluate a few tens of models to find suitable ones. As demonstrated in this paper, even a medium tier computing rig consisting of a modern, medium-range processor and a general purpose GPU can find good models in less than 5 minutes depending on the dataset. This is achievable mainly due to the partial train strategy, which is demonstrated to be an efficient method for assessing the fitness of a given model.

Overall, AMS provides an easy to use, efficient and robust algorithm for finding suitable neural network models given a dataset. We believe that the method can be easily used by somebody who has a basic knowledge of programming, making it possible for non-expert machine learning practitioners to obtain out-of-the-box solutions.

Future work will consider more complex neural network architectures such as LSTM and CNNs. Techniques for assembling entire neural network pipelines will also be explored in the future as well as the inclusion of more hyperparameters in the tuning process. An analysis of what information can be used to better characterize a neural network is also left for future work. Finally, a better way of measuring distance between two neural network architectures will be explored, since this element is of high importance for applications such as visualization and evolutionary computation.

\section*{Acknowledgement}
The authors acknowledge the funding from Conacyt Project No. 285599 and a grant (11572215) from the National Natural Science Foundation of China.

\bibliographystyle{elsart-num}
\bibliography{reference_model_selection}

\begin{thebibliography}{10}
\expandafter\ifx\csname url\endcsname\relax
  \def\url#1{\texttt{#1}}\fi
\expandafter\ifx\csname urlprefix\endcsname\relax\def\urlprefix{URL }\fi

\bibitem{Hall2009}
M.~Hall, E.~Frank, G.~Holmes, B.~Pfahringer, I.~Reutemann, P.and~Witten, The
  {WEKA} data mining software: an update, ACM SIGKDD Explorations Newsletter
  11~(1) (2009) 10--18.

\bibitem{Schaul2010}
T.~Schaul, J.~Bayer, D.~Wierstra, Y.~Sun, M.~Felder, F.~Sehnke, T.~Rucksties,
  J.~Schmidhuber, Pybrain, Journal of Machine Learning Research 11 (2010)
  743--746.

\bibitem{mlib2017}
X.~Meng, J.~Bradley, B.~Yavuz, B.~Sparks, S.~Venkataraman, D.~Liu, et~al.,
  {MLlib}: Machine learning in apache spark, Journal of Machine Learning
  Research 17~(34) (2016) 1--7.

\bibitem{TensorFlow2015}
M.~Abadi, A.~Agarwal, P.~Barham, E.~Brevdo, Z.~Chen, C.~Citro, G.~Corrado,
  et~al., {TensorFlow}: Large-scale machine learning on heterogeneous systems,
  software available from tensorflow.org (2015).
\newline\urlprefix\url{http://tensorflow.org/}

\bibitem{keras2015}
C.~Francois, Keras, \url{https://github.com/fchollet/keras} (2015).

\bibitem{caffe2014}
Y.~Jia, E.~Shelhamer, J.~Donahue, S.~Karayev, J.~Long, et~al., Caffe:
  Convolutional architecture for fast feature embedding, arXiv:1408.5093
  (2014).

\bibitem{cntk2016}
F.~Seide, A.~Agarwal, {CNTK}: Microsoft's open-source deep-learning toolkit,
  in: 22nd ACM SIGKDD International Conference on Knowledge Discovery and Data
  Mining, ACM, 2016, pp. 2135--2135.

\bibitem{AutoKeras2018}
H.~Jin, Q.~Song, X.~Hu, Auto-keras: An efficient neural architecture search
  system, arXiv:1806.10282 (2018).

\bibitem{Real2018}
E.~Real, A.~Aggarwal, Y.~Huang, Q.~V. Le, Regularized evolution for image
  classifier architecture search, arXiv:1802.01548 (2018).

\bibitem{sparks2015}
E.~Sparks, A.~Talwalkar, V.~Smith, J.~Kottalam, X.~Pan, J.~Gonzales, Automated
  model search for large scale machine learning, in: SoCC, 2015, pp. 368--380.

\bibitem{Zoph2016}
B.~Zoph, Q.~V. Le, Neural architecture search with reinforcement learning
  (2016).

\bibitem{Baker2016}
B.~Baker, O.~Gupta, N.~Naik, R.~Raskar, Designing neural network architectures
  using reinforcement learning, arXiv:1611.02167 (2016).

\bibitem{Zhong2017}
Z.~Zhong, J.~Yan, W.~Wu, J.~Shao, C.-L. Liu, Practical block-wise neural
  network architecture generation, arXiv:1708.05552 (2017).

\bibitem{Liu2018}
C.~Liu, B.~Zoph, M.~Neumann, J.~Shlens, W.~Hua, et~al., Progressive neural
  architecture search, arXiv:1712.00559 (2017).

\bibitem{Liang2017}
R.~Miikkulainen, J.~Liang, E.~Meyerson, A.~Rawal, D.~Fink, et~al., Evolving
  deep neural networks, arXiv:1703.00548 (2017).

\bibitem{Angeline1994}
P.~Angeline, G.~Saunders, J.~Pollack, An evolutionary algorithm that constructs
  recurrent neural networks, IEEE Transactions on Neural Networks 5~(1) (1994)
  54--65.

\bibitem{Suganuma2017}
M.~Suganuma, S.~Shirakawa, T.~Nagao, A genetic programming approach to
  designing convolutional neural network architectures, in: Proceedings of the
  Genetic and Evolutionary Computation Conference, GECCO '17, ACM, 2017, pp.
  497--504.

\bibitem{Thornton2016}
L.~Kotthoff, C.~Thornton, H.~H. Hoos, F.~Hutter, K.~Leyton-Brown, {Auto-WEKA
  2.0}: Automatic model selection and hyperparameter optimization in weka,
  Journal of Machine Learning Research 18~(25) (2017) 1--5.

\bibitem{Brochu2010}
E.~Brochu, V.~M. Cora, N.~de~Freitas, A tutorial on {B}ayesian optimization of
  expensive cost functions, with application to active user modeling and
  hierarchical reinforcement learning, arXiv:1012.2599 (2010).

\bibitem{Hutter2011}
F.~Hutter, H.~Hoos, K.~Leyton-Brown, Sequential model-based optimization for
  general algorithm configuration, in: Proceedings of the 5th International
  Conference on Learning and Intelligent Optimization, LION'05,
  Springer-Verlag, 2011, pp. 507--523.

\bibitem{Feurer2015}
M.~Feurer, A.~Klein, K.~Eggensperger, J.~Springenberg, M.~Blum, F.~Hutter,
  Efficient and robust automated machine learning, in: Advances in Neural
  Information Processing Systems 28, Curran Associates, Inc., 2015, pp.
  2962--2970.

\bibitem{AutoML2017}
E.~Real, S.~Moore, A.~Selle, S.~Saxena, Y.~L. Suematsu, et~al., Large-scale
  evolution of image classifiers, arXiv:1703.01041 (2017).

\bibitem{Moritz2017}
P.~Moritz, R.~Nishihara, S.~Wang, A.~Tumanov, R.~Liaw, E.~Liang, et~al., Ray:
  {A} distributed framework for emerging ai applications, arXiv:1712.05889
  (2017).

\bibitem{Engelbrecht2007}
D.~Engelbrecht, Computational Intelligence: An Introduction, Willey, 2007.

\bibitem{imagenet_cvpr09}
J.~{Deng}, W.~{Dong}, R.~{Socher}, L.~{Li}, and, Image{N}et: A large-scale
  hierarchical image database, in: 2009 IEEE Conference on Computer Vision and
  Pattern Recognition, 2009, pp. 248--255.

\bibitem{Lipton15}
Z.~C. Lipton, J.~Berkowitz, C.~Elkan, A critical review of recurrent neural
  networks for sequence learning, arXiv:1506.00019 (2015).

\bibitem{Ebehart2007}
R.~Ebehart, Y.~Shi, Computational Intelligence, Morgan Kauffman, 2007.

\bibitem{Sumathi2010}
S.~Sumathi, P.~Surekha, Computational Intelligence Paradigms. Theory and
  Applications using MATLAB, CRC Press, 2010.

\bibitem{Krishnakumar1989}
K.~Krishnakumar, Micro-genetic algorithms for stationary and non-stationary
  function optimization, in: SPIE Proceedings: Intelligent Control and Adaptive
  Systems, 1989, pp. 289--296.

\bibitem{Hillermeier2001}
C.~Hillermeier, Nonlinear Multiobjective Optimization, Springer, 2001.

\bibitem{Laredo2019a}
D.~Laredo, Z.~Chen, O.~Schütze, J.-Q. Sun, A neural network-evolutionary
  computational framework for remaining useful life estimation of mechanical
  systems, Neural Networks 116 (2019) 178 -- 187.

\bibitem{holland1992}
J.~Holland, Adaptation in Natural and Artificial Systems, MIT Press, 1992.

\bibitem{Laredo2019}
D.~Laredo, Y.~Quin, O.~Sch\"utze, J.~Q. Sun, Automatic model selection for
  neural networks, source code, [Online] (2019).
\newline\urlprefix\url{https://github.com/dlaredo/automatic_model_selection}

\bibitem{Lecun1989}
Y.~{Lecun}, L.~{Bottou}, Y.~{Bengio}, P.~{Haffner}, Gradient-based learning
  applied to document recognition, Proceedings of the IEEE 86~(11) (1998)
  2278--2324.

\bibitem{Nocedal06}
J.~Nocedal, S.~J. Wright, Numerical Optimization, 2nd Edition, Springer, New
  York, NY, USA, 2006.

\bibitem{Simard2003}
P.~Simard, D.~Steinkraus, J.~Platt, Best practices for convolutional neural
  networks applied to visual document analysis, in: the 7th International
  Conference on Document Analysis and Recognition {(ICDAR} 2003), 2-Volume Set,
  3-6 August 2003, Edinburgh, Scotland, {UK}, 2003, pp. 958--962.

\bibitem{Ciresan2010}
D.~C. Ciresan, U.~Meier, L.~M. Gambardella, J.~Schmidhuber, Deep, big, simple
  neural nets for handwritten digit recognition, Neural Computation 22~(12)
  (2010) 3207--3220.

\bibitem{Saxena2008a}
A.~Saxena, K.~Goebel, {PHM08} challenge data set,
  https://ti.arc.nasa.gov/tech/dash/groups/pcoe/prognostic-data-repository/
  (2008).

\bibitem{Saxena2008}
A.~Saxena, K.~Goebel, D.~Simon, N.~Eklund, Damage propagation modeling for
  aircraft engine run-to-failure simulation, in: International Conference On
  Prognostics and Health Management, IEEE, 2008, pp. 1--9.

\bibitem{Laredo2018}
D.~Laredo, X.~Chen, O.~Sch\"utze, J.~Q. Sun, {ANN-EA} for {RUL} estimation,
  source code, [Online] (2018).
\newline\urlprefix\url{https://github.com/dlaredo/NASA_RUL_-CMAPS-}

\bibitem{Li2018}
X.~Li, Q.~Ding, J.~Sun, Remaining useful life estimation in prognostics using
  deep convolution neural networks, Reliability Engineering and System Safety
  172 (2018) 1--11.

\bibitem{Lim2016}
P.~Lim, C.~K. Goh, K.~C. Tan, A time-window neural networks based framework for
  remaining useful life estimation, in: Proceedings International Joint
  Conference on Neural Networks, 2016, pp. 1746--1753.

\bibitem{Zhang2016}
C.~Zhang, P.~Lim, A.~Qin, K.~Tan, Multiobjective deep belief networks ensemble
  for remaining useful life estimation in prognostics, IEEE Transactions on
  Neural Networks and Learning Systems 99 (2016) 1--13.

\bibitem{Zaharia2010}
M.~Zaharia, M.~Chowdhury, M.~J. Franklin, S.~Shenker, I.~Stoica, Spark: Cluster
  computing with working sets, in: Proceedings of the 2nd USENIX Conference on
  Hot Topics in Cloud Computing, HotCloud'10, 2010, pp. 10--10.

\end{thebibliography}

\clearpage
\appendix

\section{Generated Neural Network Models}
\label{appendix:generated_models}

\setcounter{table}{0}

\begin{table}[H]
\begin{center}
\caption{Neural network model corresponding to $S_1$.}
\label{table:neural_network_model_S1}
\vspace{12pt}
\begin{tabular}{| c | c | c | c |}
\hline
Layer type & Neurons & Activation Function & Dropout Ratio \\
\hline
Fully connected & 264 & ReLU & n/a \\
Dropout & n/a & n/a & 0.65 \\
Fully Connected & 464 & ReLU & n/a\\
Dropout & n/a & n/a & 0.35\\
Fully Connected & 872 & ReLU & n/a\\
Fully Connected & 10 & Softmax & n/a\\
\hline
\end{tabular}
\end{center}

\end{table}

\begin{table}[H]
\begin{center}
\caption{Neural network model corresponding to $S_2$.}
\label{table:neural_network_model_S2}
\vspace{12pt}
\begin{tabular}{| c | c | c | c |}
\hline
Layer type & Neurons & Activation Function & Dropout Ratio \\
\hline
Fully connected & 56 & Sigmoid & n/a \\
Dropout & n/a & n/a & 0.25 \\
Fully Connected & 360 & Sigmoid & n/a\\
Fully Connected & 480 & Sigmoid & n/a\\
Fully Connected & 80 & Sigmoid & n/a\\
Dropout & n/a & n/a & 0.20\\
Fully Connected & 10 & Softmax & n/a\\
\hline
\end{tabular}
\end{center}

\end{table}

\begin{table}[H]
\begin{center}
\caption{Neural network model corresponding to $S_3$.}
\label{table:neural_network_model_S3}
\vspace{12pt}
\begin{tabular}{| c | c | c | c |}
\hline
Layer type & Neurons & Activation Function & Dropout Ratio \\
\hline
Fully connected & 264 & ReLU & n/a \\
Fully Connected & 360 & ReLU & n/a\\
Fully Connected & 480 & ReLU & n/a\\
Fully Connected & 88 & ReLU & n/a\\
Fully Connected & 872 & ReLU & n/a\\
Fully Connected & 10 & Softmax & n/a\\
\hline
\end{tabular}
\end{center}
\end{table}

\pagebreak

\end{document}